\documentclass[11pt,a4paper]{article}
\usepackage{times,latexsym}
\usepackage{url}
\usepackage[T1]{fontenc}
\usepackage{graphicx}
\usepackage{url}
\usepackage{multirow}
\usepackage{comment}
\usepackage{amssymb}
\usepackage{pifont}% http://ctan.org/pkg/pifont

\usepackage{algorithm,algcompatible,amsmath}
\usepackage{mathtools}
\usepackage{xcolor}
\usepackage{enumitem}
\usepackage{marvosym}
\usepackage{textcomp}
\usepackage{array}
\usepackage{booktabs}

\usepackage[acceptedWithA]{tacl2018v2}
\usepackage{xspace,mfirstuc,tabulary}

\newcommand{\xmark}{\ding{55}}%

\newcommand{\possessivecite}[1]{\citeauthor{#1}'s \citeyearpar{#1}}

\newif\iftaclinstructions
\taclinstructionsfalse % AUTHORS: do NOT set this to true
\iftaclinstructions
\renewcommand{\confidential}{}
\renewcommand{\anonsubtext}{(No author info supplied here, for consistency with
TACL-submission anonymization requirements)}
\newcommand{\instr}
\fi

\iftaclpubformat % this "if" is set by the choice of options

\else

\fi

\title{A Survey on Automated Fact-Checking}

\author{
 Template Author\Thanks{The {\em actual} contributors to this instruction
 document and corresponding template file are given in Section
 \ref{sec:contributors}.} \\
 Template Affiliation/Address Line 1 \\
 Template Affiliation/Address Line 2 \\
 Template Affiliation/Address Line 2 \\
  {\sf template.email@sampledomain.com} \\
}

\author{
Zhijiang Guo\thanks{$^{*}$ Equally Contributed.} \ , Michael Schlichtkrull$^{*}$, Andreas Vlachos \\
 Department of Computer Science and Technology \\
 University of Cambridge \\
  {\sf \{zg283,mss84,av308\}@cam.ac.uk} \\
}

\date{}

\begin{document}
\maketitle

\begin{abstract}
Fact-checking has become increasingly important due to the speed with which both information and misinformation can spread in the modern media ecosystem. Therefore, researchers have been exploring how fact-checking can be automated, using techniques based on natural language processing, machine learning, knowledge representation, and databases to automatically predict the veracity of claims. In this paper, we survey automated fact-checking stemming from natural language processing, and discuss its connections to related tasks and disciplines. In this process, we present an overview of existing datasets and models, aiming to unify the various definitions given and identify common concepts. Finally, we highlight challenges for future research. 
\end{abstract}

\section{Introduction}

\begin{figure}
    \centering
    \includegraphics[scale=0.71]{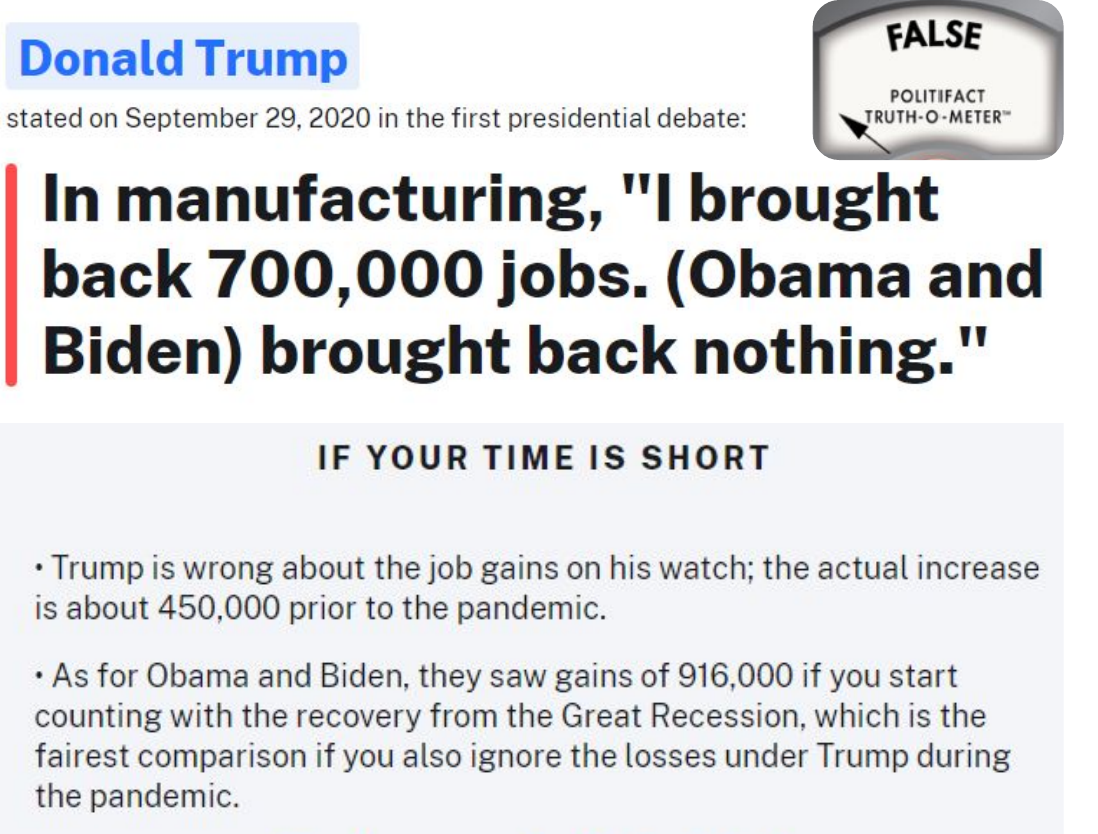}
    \vspace{-2em}
    \caption{An example of a fact-checked statement. Referring to the manufacturing sector, Donald Trump said \textit{``I brought back 700,000 jobs. Obama and Biden brought back nothing.''}  The fact-checker gave the verdict \textit{False} based on the collected evidence.}
    \label{fig:example}
    \vspace{-1em}
\end{figure}

Fact-checking is the task of assessing whether claims made in written or spoken language are true. This is an essential task in journalism, and is commonly conducted manually by dedicated organizations such as 
PolitiFact. 
%\footnote{\url{www.politifact.com}}.
% and Full Fact\footnote{\url{www.fullfact.org}}.
In addition to \textit{external} fact-checking, \textit{internal} fact-checking is also performed by publishers of newspapers, magazines, and books prior to publishing in order to promote truthful reporting.  Figure~\ref{fig:example} shows an example from PolitiFact, together with the evidence (summarized) and the verdict.

Fact-checking is a time-consuming task. To assess the claim in Figure~\ref{fig:example}, a journalist would need to search through potentially many sources to find job gains under Trump and Obama, evaluate the reliability of each source, and make a comparison. This process can take professional fact-checkers several hours or days~\citep{hassan2015detecting,Adair2017ProgressT}. Compounding the problem, fact-checkers often work under strict and tight deadlines, especially in the case of internal processes~\citep{borel2016chicago,godler2017journalistic}, and some studies have shown that less than half of all published articles have been subject to verification~\citep{lewis2008quality}. Given the amount of new information that appears and the speed with which it spreads, manual validation is insufficient.

\begin{figure*}
    \centering
    \includegraphics[scale=0.55]{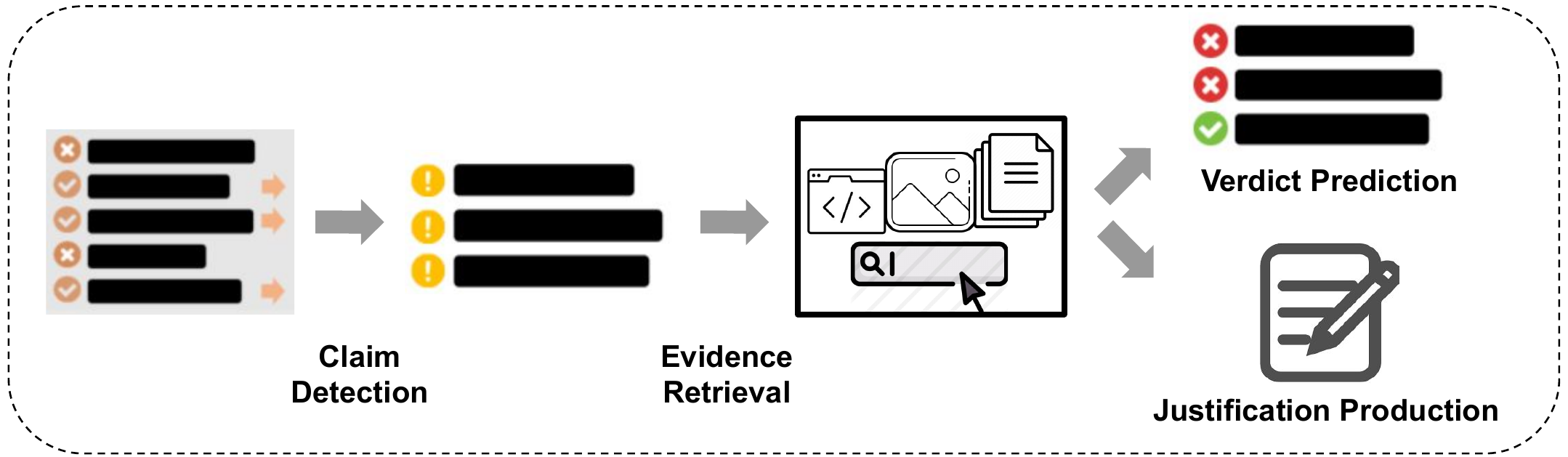}
    \vspace{-1.2em}
    \caption{A natural language processing framework for automated fact-checking.}
    \label{fig:framework}
    \vspace{-1.4em}
\end{figure*}

%Because of that, 
Automating the fact-checking process has been discussed in the context of computational journalism~\citep{Flew2010THEPO,Cohen2011ComputationalJA,2018graves}, and has received significant attention in the artificial intelligence community. 
\citet{Vlachos2014FactCT} proposed structuring it as a sequence of components -- identifying claims to be checked, finding appropriate evidence, producing verdicts --
that can be modelled as natural language processing (NLP) tasks. This motivated the development of automated pipelines consisting of subtasks that can be mapped to tasks well-explored in the NLP community. Advances were made possible by the development of datasets, consisting of either claims collected from fact-checking websites, e.g.\ Liar~\citep{Wang2017LiarLP}, or purpose-made for research, e.g.\ FEVER~\citep{Thorne2018FEVERAL}. 

A growing body of research is exploring the various tasks and subtasks necessary for the automation of fact-checking, and to meet the need for new methods to address emerging challenges. Early developments were surveyed in \citet{Thorne2018AutomatedFC}, which remains the closest to an exhaustive overview of the subject. However, their proposed framework does not include work on determining \textit{which} claims to verify (i.e.~claim detection), nor does their survey include the recent work on producing explainable, convincing verdicts (i.e.~justification production). 

Several recent papers have surveyed research focusing on individual components of the task. \citet{ZubiagaABLP18} and \citet{IslamLWX20} focus on  identifying rumours on social media, \citet{Kk2020StanceD} and \citet{Hardalov2021ASO} on detecting the stance of a given piece of evidence towards a claim, and \citet{Kotonya2020ExplainableAF} on producing explanations and justifications for fact-checks.
Finally \citet{nakov2021} surveyed automated approaches to assist fact-checking by humans.
While these surveys are extremely useful in understanding various aspects of fact-checking technology, they are fragmented and focused on specific subtasks and components; our aim is to give a comprehensive and exhaustive birds-eye view of the subject as a whole.

A number of papers have surveyed related tasks. \citet{lazer2018science} and \citet{Zhou20} surveyed work on fake news, including descriptive work on the problem, as well as work seeking to counteract fake news through computational means. A comprehensive review of NLP approaches to fake news detection was also provided in \citet{Oshikawa2020ASO}. However, fake news detection differs in scope from fact-checking, as the former focuses on %only on % currently
%actual news items
assessing news articles, and includes labelling items based on aspects not related to veracity, such as satire detection~\citep{Oshikawa2020ASO, Zhou20}. Furthermore, other factors -- such as the audience reached by the claim, and the intentions and forms of the claim -- are often considered. These factors also feature in the context of propaganda detection, recently surveyed by \citet{Martino2020ASO}. 
Unlike these efforts, the works discussed in this survey concentrate on assessing veracity of general-domain claims. Finally, \citet{shu2017fake} and \citet{SilvaVG19} surveyed research on fake news detection and fact-checking with a focus on social media data, while this survey covers fact-checking across domains and sources, including newswire, science, etc. %while \citet{Martino2020ASO} surveyed work on propaganda detection.

In this survey, we present a comprehensive and up-to-date survey of automated fact-checking, unifying various %components and 
definitions developed in previous research into a common framework. We begin by defining the three stages of our fact-checking framework -- claim detection, evidence retrieval, and claim verification, the latter consisting of verdict prediction and justification production. We then give an overview of the existing datasets and modelling strategies, taxonomizing these and contextualizing them with respect to our framework. We finally discuss key research challenges that have been addressed, and give directions for challenges which we believe should be tackled by future research. We accompany the survey with a repository,\footnote{\url{www.github.com/Cartus/Automated-Fact-Checking-Resources}}
%\footnote{\url{www.github.com/****}}
 which lists the
%provides
resources mentioned in our survey.
%and timely updates on research developments.

\section{Task Definition}
\label{sec:definition}

Figure~\ref{fig:framework} shows a NLP framework for automated fact-checking consisting of three stages:
(i) \textit{claim detection} to identify claims that require verification; (ii) \textit{evidence retrieval} to find sources supporting or refuting the claim; (iii) \textit{claim verification} to assess the veracity of the claim based on the retrieved evidence. % In the literature, 
Evidence retrieval and claim verification are sometimes tackled as a single task referred to as \textit{factual verification}, while claim detection is often tackled separately. Claim verification can be decomposed
%understood as consisting of 
into two parts that can be tackled separately or jointly: \textit{verdict prediction}, where claims are assigned truthfulness labels, and \textit{justification production}, where explanations for verdicts must be produced.

\subsection{Claim Detection}
\label{section:definitions_claim_detection}
The first stage in automated fact-checking is claim detection, where claims are selected
for verification% are identified
. Commonly, detection relies %Some previous works rely on definitions revolving 
%around 
on the concept of check-worthiness. \citet{hassan2015detecting} defined check-worthy claims as those for which the general public would be interested in knowing the truth. For example, \textit{``over six million Americans had COVID-19 in January''} would be check-worthy, as opposed to \textit{``water is wet''}.  %would not be, as the general public only have an interest in learning the veracity of the first claim.
This can involve a binary decision for each potential claim, or an importance-ranking of claims~\citep{Atanasova2018OverviewOT, BarrnCedeo2020OverviewOC}. The latter parallels standard practice in internal journalistic fact-checking, where deadlines often require fact-checkers to employ a triage system~\citep{borel2016chicago}.

Another instantiation of claim detection based on check-worthiness is rumour detection. A rumour can be defined as an unverified story or statement circulating (typically on social media)% whose truth value is unverified%at the time of posting
~\citep{Ma2016DetectingRF,ZubiagaABLP18}. Rumour detection considers language subjectivity and growth of readership through a social network~\citep{Qazvinian2011RumorHI}. Typical input to a rumour detection system is a stream of social media posts, whereupon a binary classifier has to determine if each post is rumourous. Metadata, such as the number of likes and re-posts, is often used as features to identify rumours~\citep{zubiaga2016analysing,GorrellABDKLZ19,zhang2021mining}. 

Check-worthiness and rumourousness can be subjective. For example, the importance placed on countering COVID-19 misinformation is not uniform across every social group. 
The check-worthiness of each claim also varies over time, as countering misinformation related to current events is in many cases understood to be more important than countering older misinformation (e.g.\ misinformation about COVID-19 has a greater societal impact in 2021 than misinformation about the Spanish flu). Furthermore, older rumours may have already been debunked by journalists, reducing their impact.  %the misinformation. % or awareness-raising campaigns, leading to a greater percentage of the population already knowing the truth. %The check-worthiness assigned to each claim depends on the audience, and it is not always clear which audience systems target. 
Misinformation that is harmful to marginalized communities may also be judged to be less check-worthy by the general public than misinformation that targets the majority.
%The degree of check-worthiness assigned to each claim by the general public (or any particular subset thereof, such as a set of annotators) are therefore not necessarily in alignment with the goals of system designers.
Conversely, claims \textit{originating from} marginalised groups may be subject to greater scrutiny than claims originating from the majority; for example, journalists have been shown to assign greater trust and therefore lower need for verification to stories produced by male sources % compared to other genders
~\citep{Barnoy2019}.
 Such biases could be replicated in datasets that capture the (often implicit) decisions made by journalists about which claims to prioritize. 

Instead of using subjective concepts, \citet{konstantinovskiy2018towards} framed claim detection as whether a claim makes an assertion about the world that is checkable, i.e.\ whether it is verifiable with readily available evidence. Claims based on personal experiences or opinions are uncheckable. For example, \textit{``I woke up at 7 am today''} is not checkable because appropriate evidence cannot be collected; \textit{``cubist art is beautiful''} is not checkable because it is a subjective statement.

\subsection{Evidence Retrieval}
Evidence retrieval aims to find information beyond the claim %external evidence (i.e.~beyond the claim)
-- e.g.~text, tables, knowledge bases, images, relevant metadata -- to indicate veracity% of a claim
. Some earlier efforts do not use any evidence beyond the claim itself% when predicting veracity
~\citep{Wang2017LiarLP,Rashkin2017TruthOV, Volkova2017SeparatingFF,Dungs2018CanRS}. Relying on surface patterns of claims without considering the state of the world fails to identify well-presented misinformation, including machine-generated claims~\citep{Schuster2020TheLO}. Recent developments in natural language generation have exacerbated this issue~\citep{radford2019language,Brown2020LanguageMA}, with machine-generated text sometimes being perceived % by humans 
as more trustworthy than human-written text~\citep{Zellers2019DefendingAN}. In addition to enabling verification, evidence is essential for generating verdict justifications to convince users of fact-checks.
%well-presented evidence is important to convince users of the validity of fact-checking judgments. 
%Debunking purely by calling something \textit{false} often fails to be persuasive, or in the worst case can induce a ``backfire''-effect where belief in the erroneous claim is reinforced~\citep{Lewandowsky2012}.

\textit{Stance detection} can be viewed as an instantiation of evidence retrieval, which typically assumes a more limited amount of potential evidence and predicts its stance towards the claim. For example, \citet{Ferreira2016EmergentAN} used news article headlines from the Emergent project\footnote{\url{www.cjr.org/tow_center_reports/craig_silverman_lies_damn_lies_viral_content.php}} as evidence to predict whether articles supported, refuted or merely reported a claim. The Fake News Challenge~\citep{pomerleau2017fake} further used entire documents, allowing for evidence from multiple sentences. More recently, \citet{Hanselowski2019ARA} filtered out irrelevant sentences in the summaries of fact-checking articles to obtain fine-grained evidence via stance detection. 
%
%We note that a
%Although both are classification tasks taking as input claims and evidence, stance detection differs strongly from claim verification. 
While both stance detection and evidence retrieval in the context of claim verification are classification tasks, what is considered evidence in the former %the evidence considered in the former
%attempts to determine whether a given piece of evidence  expresses a supporting or refuting view of the claim; this could
is broader, including for example a social media post responding \textit{``@AJENews @germanwings yes indeed :-(.''} to a claim% by AJENews
~\citep{GorrellABDKLZ19}. %Such evidence would not be considered appropriate for claim verification.
%Claim verification attempts to infer the truth value of the claim based on the evidence, and as such assumes a stricter definition of ``support'' and ``refute''.

A fundamental issue is that not all available information is trustworthy. Most fact-checking approaches implicitly assume access to a trusted information source such as encyclopedias (e.g.\ Wikipedia~\citep{Thorne2018FEVERAL}) or results provided (and thus vetted) by search engines~\citep{Augenstein2019MultiFCAR}. \textit{Evidence} is then defined as information that can be retrieved from this source, and \textit{veracity} as coherence with the evidence. For real-world applications, evidence must be curated through the manual efforts of journalists~\citep{borel2016chicago}, automated means~\citep{li2016survey}, or their combination. 
For example, Full Fact uses tables and legal documents from government organisations as evidence.\footnote{\url{www.fullfact.org/about/frequently-asked-questions}}

\subsection{Verdict Prediction}
Given an identified claim and the pieces of evidence retrieved for it, verdict prediction attempts to determine the veracity of the claim. The simplest approach is binary classification, e.g.~labelling a claim as true or false~\citep{Nakashole2014LanguageAwareTA, Popat2016CredibilityAO, Potthast2018ASI}. When evidence is used to verify the claim, it is often preferable to use supported/refuted (by evidence) instead of true/false respectively, as in many cases the evidence itself is not assessed by the systems. More broadly it would be dangerous to make such strong claims about the world given the well-known limitations~\citep{2018graves}.

% Ordinal verdicts
Many versions of the task employ finer-grained classification schemes. A simple extension is to use an additional label denoting a lack of % that the evidence does not contain 
information to predict %make a decision on 
the veracity of the claim~\citep{Thorne2018FEVERAL}. Beyond that, some datasets and systems follow the approach taken by journalistic fact-checking agencies, employing multi-class labels representing degrees of truthfulness~\citep{Wang2017LiarLP,Alhindi2018WhereIY, shahifakecovid, Augenstein2019MultiFCAR}. %Commonly, these categories are modelled as nominal classification; however, it has been suggested that degree of truthfulness should be modelled as an ordinal classification task~\citep{Vlachos2014FactCT}. 
%We note that employing fine-grained truthfulness labels in a meaningful way can be challenging. The use of such labels by journalists has come under criticism~\citep{uscinski2013epistemology}, as in-between labels like \textit{``mostly true''} often represent ``meta-ratings'' for composite claims consisting of multiple elementary claims of different veracity, and rely on highly subjective judgments about how to weigh individual components in the final rating.

\begin{table*}
\centering
\scalebox{0.74}{
\begin{tabular}{lccccccc}
\toprule
Dataset & Type & Input & \#Inputs & Evidence & Verdict & Sources & Lang \\
\midrule
CredBank~\citep{MitraG15} & Worthy & Aggregate & 1,049  & Meta & 5 Classes & Twitter & En \\
Weibo~\citep{Ma2016DetectingRF} & Worthy & Aggregate & 5,656  & Meta & 2 Classes &  Twitter/Weibo & En/Ch  \\
% Weibo20 \citep{zhang2021mining} & Aggregate & 6,362 & 2 & Meta & Weibo & Ch \\
PHEME~\citep{zubiaga2016analysing} & Worthy & Individual & 330  & Text/Meta & 3 Classes & Twitter & En/De \\
% RumourEval \citep{Derczynski2017SemEval2017T8} & Worthy & Individual & 325  & Text/Meta & 3 & Twitter & En \\
RumourEval19~\citep{GorrellABDKLZ19} & Worthy & Individual & 446 & Text/Meta & 3 Classes & Twitter/Reddit & En \\
DAST~\citep{Lillie2019JointRS} & Worthy & Individual & 220  & Text/Meta & 3 Classes & Reddit & Da \\
Suspicious~\citep{Volkova2017SeparatingFF} & Worthy & Individual & 131,584  & \xmark & 2/5 Classes & Twitter & En \\
CheckThat20-T1~\citep{BarrnCedeo2020OverviewOC} & Worthy & Individual & 8,812  & \xmark & Ranking & Twitter & En/Ar \\
CheckThat21-T1A~\citep{NakovMEBMSAHHBN21} & Worthy & Individual & 17,282  & \xmark & 2 Classes & Twitter & Many \\
% FauxBuster \citep{Zhang2018FauxBusterAC} & Worthy & Individual$^{\dagger}$ & 917  & Meta  & 2 & Twitter/Reddit & En \\
% COVIDLIES \citep{Hossain2020COVIDLiesDC} & Worthy & Individual & 6,761 & Misconception & 3 & Twitter & En \\
% \midrule
Debate~\citep{hassan2015detecting} & Worthy & Statement & 1,571  & \xmark & 3 Classes & Transcript & En \\
ClaimRank~\citep{Gencheva2017ACA} & Worthy & Statement & 5,415  & \xmark & Ranking & Transcript & En \\
CheckThat18-T1~\citep{Atanasova2018OverviewOT} & Worthy & Statement & 16,200  & \xmark & Ranking & Transcript & En/Ar \\
% CheckThat19-T1 \citep{CheckThat2019} & Claim & 23,500 & ranking & \xmark & Transcript & En \\
% CheckThat20-T5 \citep{BarrnCedeo2020OverviewOC} & Claim & 64,290 & ranking & \xmark & Transcript & En \\
% CheckThat21-T1B \citep{clef2021} & Claim & 84,290 & ranking & \xmark & Transcript & En \\
\midrule
%\midrule
CitationReason~\citep{redi2019}  & Checkable & Statement & 4,000  & Meta & 13 Classes & Wikipedia & En \\
PolitiTV~\citep{konstantinovskiy2018towards} & Checkable & Statement & 6,304  & \xmark & 7 Classes & Transcript & En \\
%SemEval19-TA\citep{Mihaylova2019SemEval2019T8} & Checkable & Question & 2,310 & Meta  & 3 & Forum & En \\
\bottomrule
\end{tabular}}
\vspace{-0.5em}
\caption{Summary of claim detection datasets. Input can be a set of posts (aggregate) or an individual post from social media, or a statement. Evidence include text and metadata. Verdict can be a multi-class label or a rank list.}
% En, De, Ch, Ar, and Da indicate English, German, Chinese, Arabic, and Danish, respectively.}
% \citet{NakovMEBMSAHHBN21} offered datasets in Arabic, Bulgarian, English, and Spanish.}
\label{tab:detection}
\vspace{-1em}
\end{table*}

\subsection{Justification Production}
\label{section:definitions_justifications}

% Intro
Justifying decisions is an important part of journalistic fact-checking, as fact-checkers need to convince readers of their interpretation of the evidence~\citep{uscinski2013epistemology,borel2016chicago}. Debunking purely by calling something \textit{false} often fails to be persuasive, and %or in the worst case 
can induce a ``backfire''-effect where belief in the erroneous claim is reinforced~\citep{Lewandowsky2012}. This need is even greater for automated fact-checking, which may employ black-box components. When developers deploy black-box models whose decision-making processes cannot be understood, these artefacts can lead to unintended, harmful consequences~\citep{oneil2016weapons}. Developing techniques that explain model predictions has been suggested as a potential remedy to this problem~\citep{Lipton18},
%
%The potentially sensitive applications of fact verification technology heighten the need for users and developers to understand how and why predictions are made. 
and recent work has focused on %extending the claim verification task to also include 
the generation of \textit{justifications} (see \possessivecite{Kotonya2020ExplainableAF} survey of explainable claim verification). Research so far has focused on justification production for claim verification, as the latter is often the most scrutinized stage in fact-checking. Nevertheless, explainability may also be desirable and necessary
for the other stages in our framework. % of Figure~\ref{fig:framework}.%, e.g.\ for claim identification which can rely on check-worthiness assessments.

% Short description of how modelled
Justification production for claim verification typically relies on one of four strategies. First, attention weights can be used to highlight the salient parts of the evidence, in which case justifications typically consist of scores for each evidence token~\citep{Popat2018DeClarEDF, Shu2019dEFENDEF, Lu2020GCANGC}. Second, decision-making processes can be designed to be understandable by human experts, e.g.\ by relying on logic-based systems~\citep{GadElrab2019ExFaKTAF, AhmadiLPS19}; in this case, the justification is typically the derivation for the veracity of the claim. Finally, the task can be modelled as a form of summarization, where systems generate textual explanations for their decisions~\citep{Atanasova2020GeneratingFC}. 
While some of these justification types require additional components, we did not introduce a fourth stage in our framework as in some cases the decision-making process of the model
%process with which the model reaches the verdict 
is self-explanatory~\citep{GadElrab2019ExFaKTAF, AhmadiLPS19}. 

% Difference between explanations and evidence
%A necessary -- but not sufficient -- part of a 
A basic form of justification is to show which pieces of evidence were used to reach a verdict.
%, as is required in FEVER-style tasks~\citep{Thorne2018FEVERAL}. 
However, a justification must also explain \textit{how} the retrieved evidence was used, explain any assumptions or commonsense facts employed, and show the reasoning process taken to reach the verdict.
Presenting the  evidence returned by a 
%an evidence set through a %high-precision 
retrieval system can as such be seen as a rather weak baseline for justification production, as it does not explain the process used to reach the verdict.
There is furthermore a subtle difference between evaluation criteria for evidence and justifications: good evidence facilitates the production of a correct verdict; a good justification accurately reflects the reasoning of the model through a readable and plausible explanation, \textit{regardless} of the correctness of the verdict. This introduces different considerations for justification production, e.g.\ \textit{readability} (how accessible an explanation is to humans), \textit{plausibility} (how convincing an explanation is), and \textit{faithfulness} (how accurately an explanation reflects the reasoning of the model)~\citep{jacovi2020towards}.

\section{Datasets}
\label{sec:data}

\begin{table*}
\centering
\scalebox{0.74}{
\begin{tabular}{lccccccccccc}
\toprule
Dataset & Input & \#Inputs & Evidence & Verdict & Sources & Lang \\
\midrule
CrimeVeri~\citep{bachenko-etal-2008-verification} & Statement & 275 & \xmark & 2 Classes & Crime & En \\
Politifact~\citep{Vlachos2014FactCT}  & Statement  & 106  & Text/Meta & 5 Classes & Fact Check & En \\
StatsProperties~\citep{vlachos-riedel-2015-identification} & Statement & 7,092  & KG & Numeric & Internet & En \\
% NewsTrust \citep{Mukherjee2015LeveragingJI}  & Statement  & 106 & 5 & Text  & Fact Check & En \\
Emergent~\citep{Ferreira2016EmergentAN} & Statement & 300  & Text & 3 Classes & Emergent & En \\
CreditAssess~\citep{Popat2016CredibilityAO} & Statement & 5,013 & Text & 2 Classes & Fact Check/Wiki & En \\
PunditFact~\citep{Rashkin2017TruthOV} & Statement  & 4,361  & \xmark & 2/6 Classes & Fact Check & En \\
Liar~\citep{Wang2017LiarLP} & Statement  & 12,836  & Meta & 6 Classes & Fact Check & En \\
% Liar-Plus \citep{Alhindi2018WhereIY} & Statement & 12,836 & 6 & Text/Meta & Fact Check & En \\
Verify~\citep{Baly2018IntegratingSD} & Statement & 422  & Text & 2 Classes & Fact Check & Ar/En \\
CheckThat18-T2 \citep{Barron-CedenoES18} & Statement & 150  & \xmark & 3 Classes & Transcript & En \\
% CheckThat19-T2 \citep{HasanainSEBN19} & Statement &  69  & Text & 2 & News & En\\
Snopes~\citep{Hanselowski2019ARA} & Statement & 6,422  & Text & 3 Classes & Fact Check & En \\
MultiFC~\citep{Augenstein2019MultiFCAR} & Statement & 36,534  & Text/Meta & 2-27 Classes &  Fact Check & En \\
Climate-FEVER~\citep{diggelmann2020} & Statement & 1,535 & Text  & 4 Classes & Climate & En \\
SciFact~\citep{Wadden2020FactOF} & Statement & 1,409  & Text & 3 Classes & Science & En \\
PUBHEALTH~\citep{kotonya2020explainable} & Statement & 11,832  & Text & 4 Classes & Fact Check & En\\
COVID-Fact~\citep{SaakyanCM20} & Statement & 4,086  & Text & 2 Classes & Forum & En\\
X-Fact~\citep{gupta-srikumar-2021-x} & Statement & 31,189  & Text & 7 Classes & Fact Check & Many\\
\midrule
cQA~\citep{Mihaylova2018FactCI} & Answer & 422  & Meta & 2 Classes & Forum & En \\
% SemEval19-TB \citep{Mihaylova2019SemEval2019T8} & Answer & 917  & Meta & 3 & Forum & En \\
AnswerFact~\citep{Zhang2020AnswerFactFC} & Answer & 60,864 & Text & 5 Classes & Amazon & En \\
\midrule
% Hoax \citep{TacchiniBVMA17} & Article & 15,500  & Meta & 2 Classes & Facebook & En \\
NELA~\citep{HorneKA18} & Article & 136,000  & \xmark & 2 Classes & News & En \\
BuzzfeedNews \citep{Potthast2018ASI} & Article & 1,627 & Meta & 4 Classes & Facebook & En \\
BuzzFace~\citep{SantiaW18} & Article & 2,263 & Meta & 4 Classes & Facebook & En \\
% NELA18 \citep{Nrregaard2019NELAGT2018AL} & Article & 713,000 & 2 & \xmark & News & En \\
% NELA19 \citep{gruppi2020nela} & Article & 1.12M & 2 & \xmark & News & En \\
FA-KES~\citep{SalemFEJF19} & Article & 804 & \xmark & 2 Classes & VDC & En \\
FakeNewsNet \citep{Shu2018FakeNewsNetAD} & Article & 23,196  & Meta & 2 Classes & Fact Check & En \\
%FakeHealth \citep{DaiSW20} & Article & 2,296  & Meta & 2 Classes & Health & En \\
FakeCovid~\citep{shahifakecovid} & Article & 5,182  & \xmark & 2 Classes & Fact Check & Many \\
% \midrule
% Fauxtography \citep{Zlatkova2019FactCheckingMF} & Statement/Image & 1,233  & Meta & 2 & Fact Check/News & En \\
% Fakeddit \citep{Nakamura2020rFakedditAN} & Statement/Image & 1.06M  & Meta & 2,3,6 & Reddit & En \\
\bottomrule
\end{tabular}}
\vspace{-0.5em}
\caption{Summary of factual verification datasets with natural inputs. KG denotes knowledge graphs. ChectThat18 has been extended later~\citep{HasanainSEBN19,BarrnCedeo2020OverviewOC,NakovMEBMSAHHBN21}. NELA has been updated by adding more data from more diverse sources~\citep{NorregaardHA19,gruppi2020,gruppi2021}
}
\label{tab:natural}
\vspace{-1em}
\end{table*}
% Datasets can be analysed along three comparative axes aligned with the three stages of the fact-checking framework (Figure~\ref{fig:framework}): the input, the evidence used to make decisions, and the verdicts and justifications which constitute the output. In this section we bring together efforts that emerged in different communities using different terminologies, but nevertheless could be used to develop and evaluate models for the same task.

Datasets can be analysed along three axes aligned with three stages of the fact-checking framework (Figure~\ref{fig:framework}): the input, the evidence used, and verdicts and justifications which constitute the output. In this section we bring together efforts that emerged in different communities using different terminologies, but nevertheless could be used to develop and evaluate models for the same task.

\subsection{Input}
We first consider the inputs to claim detection (summarized in Table~\ref{tab:detection}) as their format and content influences the rest of the process. 
A typical input is a social media post with textual content. \citet{zubiaga2016analysing} constructed PHEME based on source tweets in English and German that sparked a high number of retweets exceeding a predefined threshold. \citet{Derczynski2017SemEval2017T8} introduced the shared task RumourEval using the English section of PHEME; for the 2019 iteration of the shared task, this dataset was further expanded to include Reddit and new Twitter posts~\citep{GorrellABDKLZ19}.  Following the same annotation strategy, \citet{Lillie2019JointRS} constructed a Danish dataset by collecting posts from Reddit. Instead of considering only source tweets, subtasks in CheckThat~\citep{BarrnCedeo2020OverviewOC,NakovMEBMSAHHBN21} viewed every post as part of the input. A set of auxiliary questions, such as \textit{``does it contain a factual claim?''}, \textit{``is it of general interest?''}, were created to help annotators identify check-worthy posts. Since an individual post may contain limited context, other works~\citep{MitraG15,Ma2016DetectingRF,zhang2021mining} represented each claim by a set of relevant posts, e.g.\ the thread they originate from. 
 
The second type of textual input is a document consisting of multiple claims. For Debate~\citep{hassan2015detecting}, professionals were asked to select check-worthy claims from U.S.\ presidential debates to ensure good agreement and shared understanding of the assumptions. On the other hand, \citet{konstantinovskiy2018towards} collected checkable claims from transcripts by crowd-sourcing, where workers labelled claims based on a predefined taxonomy. Different from prior works focused on the political domain, \citet{redi2019} sampled sentences that contain citations from Wikipedia articles, and asked crowd-workers to annotate them based on citation policies.

Next, we discuss the inputs to factual verification. The most popular type of input to verification is textual claims, which is expected given they are often the output of claim detection. These tend to be sentence-level statements, which is a practice common among fact-checkers in order to include only the context relevant to the claim~\citep{mena2019principles}. Many existing efforts~\citep{Vlachos2014FactCT, Wang2017LiarLP, Hanselowski2019ARA, Augenstein2019MultiFCAR} constructed datasets by crawling real-world claims from dedicated websites (e.g.\ Politifact) due to their availability (see Table~\ref{tab:natural}). Unlike previous work that focus on English, \citet{gupta-srikumar-2021-x} collected non-English claims from 25 languages.

Others extract claims from specific domains, such as science~\citep{Wadden2020FactOF}, climate~\cite{diggelmann2020}, and public health~\citep{kotonya2020explainable}. Alternative forms of sentence-level inputs, such as answers from question answering forums, have also been considered~\citep{Mihaylova2018FactCI,Zhang2020AnswerFactFC}. There have been approaches that consider a passage~\citep{mihalcea2009lie,PrezRosas2018AutomaticDO} or an entire article~\citep{HorneKA18,SantiaW18,Shu2018FakeNewsNetAD} %,DaiSW20}
as input. However, the implicit assumption that every claim in it is either factually correct or incorrect is problematic, and thus rarely practised by human fact-checkers~\citep{uscinski2013epistemology}. 

\begin{table*}
\centering
\scalebox{0.74}{
\begin{tabular}{lccccccccccc}
\toprule
Dataset & Input & \#Inputs & Evidence & Verdict & Sources & Lang \\
\midrule
KLinker~\citep{Ciampaglia2015ComputationalFC} & Triple & 10,000  & KG & 2 Classes & Google/Wiki & En \\
PredPath~\citep{ShiW16} & Triple & 3,559  & KG & 2 Classes & Google/Wiki & En \\
% KStream~\citep{Shiralkar2017FindingSI} & Triple & 18,431  & KG & 2 Classes & Google/Wiki/WSDM & En \\
KStream~\citep{Shiralkar2017FindingSI} & Triple & 18,431  & KG & 2 Classes & Google/Wiki & En \\
UFC~\citep{kim-choi-2020-unsupervised} & Triple & 1,759  & KG & 2 Classes & Wiki & En \\
\midrule
LieDetect~\citep{mihalcea2009lie} & Passage & 600  & \xmark & 2 Classes & News & En \\
FakeNewsAMT~\citep{PrezRosas2018AutomaticDO} & Passage & 680  & \xmark & 2 Classes & News & En \\
\midrule
FEVER~\citep{Thorne2018FEVERAL} & Statement & 185,445  & Text & 3 Classes & Wiki & En \\
HOVER~\citep{jiang2020hover} & Statement & 26,171  & Text & 2 Classes & Wiki & En  \\
WikiFactCheck~\citep{SatheALPP20} & Statement & 124,821  & Text & 2 Classes & Wiki & En  \\
% Symmetric \citep{Schuster2019TowardsDF} & Statement & 956  & Text & 3 & Wikipedia & En \\
% FM2 \citep{foolme2021} & Statement & 12,968  & Text & 3 & Wikipedia & En \\
VitaminC~\citep{vitaminc2021} & Statement & 488,904  & Text & 3 Classes & Wiki & En \\
TabFact~\citep{Chen2020TabFactAL} & Statement & 92,283  & Table & 2 Classes & Wiki & En \\
InfoTabs~\citep{Gupta2020INFOTABSIO} & Statement & 23,738  & Table & 3 Classes & Wiki & En \\
Sem-Tab-Fact~\citep{wang2021semeval} & Statement & 5,715  & Table & 3 Classes & Wiki & En \\
FEVEROUS~\citep{Aly2021FEVEROUSFE} & Statement & 87,026  & Text/Table & 3 Classes & Wiki & En \\
ANT~\citep{Khouja2020StancePA} & Statement & 4,547  & \xmark & 3 Classes & News & Ar \\
DanFEVER~\citep{NorregaardD21} & Statement & 6,407  & Text & 3 Classes & Wiki & Da \\
\bottomrule
\end{tabular}}
\vspace{-0.5em}
\caption{Summary of factual verification datasets with artificial inputs. Google denotes Google Relation Extraction Corpora, and WSDM means the WSDM Cup 2017 Triple Scoring challenge.}
\label{tab:artificial}
\vspace{-1em}
\end{table*}

In order to better control the complexity of the task, efforts listed in Table~\ref{tab:artificial} created claims artificially. \citet{Thorne2018FEVERAL} had annotators mutate sentences from Wikipedia articles to create claims. Following the same approach, \citet{Khouja2020StancePA} and \citet{NorregaardD21} constructed Arabic and Danish datasets respectively. Another frequently considered option is subject-predicate-object triples, e.g.\ \textit{(London, city\textunderscore in, UK)}. The popularity of triples as input stems from the fact that they facilitate fact-checking against knowledge bases~\citep{Ciampaglia2015ComputationalFC,ShiW16,Shiralkar2017FindingSI,kim-choi-2020-unsupervised} such as DBpedia~\citep{AuerBKLCI07}, SemMedDB~\citep{KilicogluSFRR12}, and KBox~\citep{NamKKJHC18}. However, such approaches implicitly assume the non-trivial conversion of text into triples.

\subsection{Evidence}
A popular type of evidence often considered is metadata, such as publication date, sources, user profiles, etc. %Metadata provides insights that are useful for claim detection, for example, domain-specific metadata such as likes, or numbers of re-posts.
However, while it offers information complementary to textual sources or structural knowledge which is useful when the latter are unavailable~\citep{Wang2017LiarLP, Potthast2018ASI}, it %Nevertheless, metadata
does not provide evidence grounding the claim.

Textual sources, such as news articles, academic papers, and Wikipedia documents, are one of the most commonly used types of evidence for fact-checking. \citet{Ferreira2016EmergentAN} used the headlines of selected news articles, and~\citet{pomerleau2017fake} used the entire articles instead as the evidence for the same claims. Instead of using news articles, \citet{Alhindi2018WhereIY} and \citet{Hanselowski2019ARA} extracted summaries accompanying fact-checking articles about the claims as evidence. Documents from specialized domains such as science and public health have also been considered~\citep{Wadden2020FactOF, kotonya2020explainable,Zhang2020AnswerFactFC}.

The aforementioned works assume that evidence is given for every claim, which is not conducive to developing systems that need to retrieve evidence from a large knowledge source. Therefore, \citet{Thorne2018FEVERAL} and \citet{jiang2020hover} considered Wikipedia as the source of evidence and annotated the sentences supporting or refuting each claim. \citet{vitaminc2021} constructed VitaminC based on factual revisions to
Wikipedia, in which evidence pairs are nearly identical in language and content, with the exception that one supports a claim while the other does not. However, these efforts restricted world knowledge to a single source (Wikipedia), ignoring the challenge of retrieving evidence from heterogeneous sources on the web. To address this, other works~\citep{Popat2016CredibilityAO, Baly2018IntegratingSD, Augenstein2019MultiFCAR} retrieved evidence from the Internet, but the search results were not annotated. Thus, it is possible that irrelevant information is present in the evidence, while information that is necessary for verification is missing.

Though the majority of studies focus on unstructured evidence (i.e.\ textual sources), structured knowledge has also been used. For example, the truthfulness of a claim expressed as an edge in a knowledge base (e.g.\ DBpedia) can be predicted by the graph topology~\citep{Ciampaglia2015ComputationalFC,ShiW16,Shiralkar2017FindingSI}. However, while graph topology can be an indicator of plausibility, it does not provide conclusive evidence. A claim that is not represented by a path in the graph, or that is represented by an unlikely path, is not necessarily false. %does not have a path representing it in a graph, or that is represented by an unlikely path, is not necessarily false. 
The knowledge base approach assumes that true facts relevant to the claim are present in the graph; but given the incompleteness of even the largest knowledge bases, this is not realistic~\citep{bordes2013translating, SocherCMN13}.%pmlr-v48-trouillon16,SchlichtkrullKB18}.% One limitation of using knowledge graphs is that it assumes that the true facts relevant to the claim are present in them, but it is not possible to capture and store all related facts in the graph. 

Another type of structural knowledge is semi-structured data (e.g.\ tables), which is ubiquitous thanks to its ability to convey important information in a concise and flexible manner. 
Early work by \citet{vlachos-riedel-2015-identification} used tables extracted from Freebase \citep{BollackerEPST08} to verify claims retrieved from the web about statistics of countries such as population, inflation, etc. \citet{Chen2020TabFactAL} and \citet{Gupta2020INFOTABSIO} studied fact-checking textual claims against tables and info-boxes from Wikipedia. \citet{wang2021semeval} extracted tables from scientific articles and required evidence selection in the form of cells selected from tables. \citet{Aly2021FEVEROUSFE} further considered both text and table for factual verification, while explicitly requiring the retrieval of evidence.

%The trustworthiness of the evidence plays a significant role in manual fact-checking. For example, dedicated organizations (e.g.\ Full Fact) collect evidence by referring to vetted sources such as data table, legal documents from governmental organisations\footnote{\url{https://fullfact.org/about/frequently-asked-questions/}}. However, most existing datasets adopted a closed-world paradigm where assumes that all evidence from crowd-generated data (e.g.\ DBpedia, Wikipedia) is true to simplify the development and evaluation of automated systems.

\subsection{Verdict \& Justification}
The verdict in early efforts~\citep{bachenko-etal-2008-verification,mihalcea2009lie} is a binary label, i.e.\ \textit{true}/\textit{false}. However, fact-checkers usually employ multi-class labels to represent degrees of truthfulness (e.g.\ \textit{true}, \textit{mostly-true}, \textit{mixture}, etc),\footnote{  \url{www.snopes.com/fact-check-ratings}} which were considered by \citet{Vlachos2014FactCT} and 
\citet{Wang2017LiarLP}. 
Recently, \citet{Augenstein2019MultiFCAR} collected claims from different sources, where the number of labels vary greatly, ranging from 2 to 27. Due to the difficulty of mapping veracity labels onto the same scale, they didn't attempt to harmonize them across sources. On the other hand, other efforts~\citep{Hanselowski2019ARA,kotonya2020explainable,gupta-srikumar-2021-x} performed normalization by post-processing the labels based on rules to simplify the veracity label. For example, \citet{Hanselowski2019ARA} mapped \textit{mixture}, \textit{unproven}, and \textit{undetermined} onto \textit{not enough information}.

Unlike prior datasets that only required outputting verdicts, FEVER~\citep{Thorne2018FEVERAL} expected the output to contain both sentences forming the evidence and a label (e.g.\ support, refute, not enough information). 
% indicating whether the evidence supports, refutes, or does not contain enough information to verify the claim. 
Later datasets with both natural ~\citep{Hanselowski2019ARA,Wadden2020FactOF} and artificial claims~\citep{jiang2020hover,vitaminc2021} also adopted this scheme, where the output expected is a combination of multi-class labels and extracted evidence. 

Most existing datasets do not contain textual explanations provided by journalists as justification for verdicts. \citet{Alhindi2018WhereIY} extended the Liar dataset with summaries extracted from fact-checking articles. While originally intended as an auxiliary task to improve claim verification, these justifications have been used as explanations~\citep{Atanasova2020GeneratingFC}.
Recently, \citet{kotonya2020explainable} constructed the first dataset which explicitly includes gold explanations. These consist of fact-checking articles and other news items, %articles from news sources, 
which can be used to train natural language generation models to provide post-hoc justifications for the verdicts, 
% and further improve the performance of veracity prediction~\citep{Atanasova2020GeneratingFC}.
However, using fact-checking articles is not realistic, as they are not available during inference, which makes the trained system unable to provide justifications based on retrieved evidence. % under real-world settings.

\section{Modelling Strategies}
We now turn to surveying modelling strategies for the various components of our framework.
%, as well as the challenges they address. 
The most common approach is to build separate models for each component and apply them in pipeline fashion.
Nevertheless, joint approaches % where one model solves several component tasks 
%are also sometimes
have also been developed, either through end-to-end %multitask
learning or by modelling the joint output distributions of multiple components.

\subsection{Claim Detection}
Claim detection is typically framed as a classification task, where models predict whether claims are checkable or check-worthy. This is challenging, especially in the case of check-worthiness: rumourous and non-rumourous information is often difficult to distinguish, and the volume of claims analysed in real-world scenarios -- e.g. all posts published to a social network every day -- prohibits the retrieval and use of evidence. Early systems employed supervised classifiers with feature engineering, relying on surface features like Reddit karma and up-votes~\citep{Aker2017SimpleOS}, Twitter-specific types~\citep{Enayet2017NileTMRGAS}, named entities and verbal forms in political transcripts~\citep{Zuo2018AHR}, or lexical and syntactic features~\citep{zhou2020fakenews}.

Neural network approaches based on sequence- or graph-modelling have recently become popular, as they allow models to use the context of surrounding social media activity to inform decisions. This can be highly beneficial, as the ways in which information is discussed and shared by users are strong indicators of rumourousness~\citep{zubiaga2016analysing}. \citet{Kochkina2017TuringAS} employed an LSTM~\citep{HochreiterS97} to model branches of tweets, %processing sequences of posts and outputting a label at each time step. 
\citet{Ma2018RumorDO} used Tree-LSTMs~\citep{tai2015improved} to directly encode the structure of threads, and \citet{Guo2018RumorDW} modelled the hierarchy by using attention networks. Recent work explored fusing more domain-specific features into neural models~\citep{zhang2021mining}. Another popular approach is to use Graph Neural Networks~\citep{Kipf2017SemiSupervisedCW} to model the propagation behaviour of a potentially rumourous claim~\citep{monti2019fake,li-etal-2020-exploiting,yang2020rumor}.
%focus on the propagation behaviour of a potentially rumourous claim, using Graph Neural Networks~\citep{Kipf2017SemiSupervisedCW} to model its spread through social networks~\citep{monti2019fake,li-etal-2020-exploiting,yang2020rumor}.

Some works tackle claim detection and claim verification jointly, labelling potential claims as \textit{true rumours}, \textit{false rumours}, or \textit{non-rumours}~\citep{code2017automatically, Ma2018RumorDO}. This allows systems to exploit specific features useful for both tasks, such as the different spreading patterns of false and true rumours~\citep{zubiaga2016analysing}. %We note, though, that
%However, the v 
Veracity predictions made by such systems are to be considered preliminary, as they are made without evidence.

% Main challenges: Scale, how to classify w/o retrieving evidence (b/c prohibitive scale), subjectivity, multiple modalities (see ACL paper on meme propaganda)

\subsection{Evidence Retrieval \& Claim Verification}
As mentioned in Section~\ref{sec:definition}, evidence retrieval and claim verification are commonly addressed together. Systems mostly operate as a pipeline consisting of an evidence retrieval module and a verification module~\citep{Thorne18Fact}, but there are exceptions where these two modules are trained jointly~\citep{Riloff18}. 

Claim verification can be seen as a form of Recognizing Textual Entailment (RTE; \citealt{DaganDMR10,Bowman2015ALA}), predicting whether the evidence supports or refutes the claim. 
%The primary challenge for retrieval systems is scale -- the number of articles in potential information sources such as Wikipedia, LexisNexis, or Google News is massive. Retrieval modules are therefore very limited in how much processing can be spent on each potential piece of evidence. This challenge has been addressed through  various
Typical retrieval strategies include commercial search APIs, Lucene indices, entity linking, or ranking functions like dot-products of TF-IDF vectors~\citep{Thorne18Fact}. Recently, dense retrievers employing learned representations and fast dot-product indexing~\citep{johnson2017faiss} have shown strong performance~\citep{lewis2020retrieval, maillard-etal-2021-multi}. %Efficient retrieval often comes as the expense of precision. To alleviate this,
To improve precision% of generic retrieval approaches
,
more complex models -- for example stance detection systems -- can be deployed as second, fine-grained filters to re-rank retrieved evidence~\citep{Thorne18Fact, nie-etal-2019-revealing, Nie2019CombiningFE, Hanselowski2019ARA}. Similarly, evidence can be re-ranked implicitly during verification in late-fusion systems
%late-fusion strategies where evidence is re-ranked implicitly as part of the verification system have also been proposed
~\citep{ma-etal-2019-sentence, schlichtkrull2020}.
An alternative approach was proposed by \citet{Fan2020GeneratingFC}, who retrieved evidence using question generation and question answering via search engine results. 
Some work avoids retrieval by making a \textit{closed-domain assumption} and %, i.e.\ assuming access to an oracle retriever, and 
evaluating in a setting where appropriate evidence has already been found~\citep{Ferreira2016EmergentAN,Chen2020TabFactAL, zhong-etal-2020-logicalfactchecker, yang-etal-2020-program, eisenschlos-etal-2020-understanding}; this, however, is unrealistic. Finally, \citet{allein2021time} took into account the timestamp of the evidence in order to improve veracity prediction accuracy.

% What to do with evidence
If only a single evidence document is retrieved, verification can be directly modelled as RTE. However, both real-world claims~\citep{Augenstein2019MultiFCAR, Hanselowski2019ARA, kotonya2020explainable}, as well as those created for research purposes~\citep{Thorne2018FEVERAL,jiang2020hover,vitaminc2021} often require reasoning over and combining multiple pieces of evidence.  %This introduces significant challenges compared to considering a single piece of evidence. 
A simple approach is to treat multiple pieces of evidence as one by concatenating them into a single string~\citep{luken-etal-2018-qed, Nie2019CombiningFE}, and then employ a textual entailment model to infer whether the evidence supports or refutes the claim. More recent systems employ specialized components to aggregate multiple pieces of evidence. This allows the verification of more complex claims where several pieces of information must be combined, and addresses the case where the retrieval module returns several highly-related documents all of which \textit{could} (but might not) contain the right evidence~\citep{yoneda-etal-2018-ucl,Zhou2019GEARGE, ma-etal-2019-sentence,Liu2020KernelGA, Zhong2020ReasoningOS, schlichtkrull2020}.

Some early work does not include evidence retrieval at all, performing verification purely on the basis of surface forms and metadata~\citep{Wang2017LiarLP,Rashkin2017TruthOV,Dungs2018CanRS}. Recently \citet{lee-etal-2020-language} 
considered using the information stored in the weights of a large pretrained language model -- BERT~\citep{Devlin2019BERTPO} -- as the only source of evidence, as it has been shown competitive in knowledge base completion \citep{petroni-etal-2019-language}.
Without explicitly considering evidence such approaches are likely to propagate biases learned during training, and render justification production impossible~\citep{lee-etal-2021-towards,PanCXKW20}.
%pecial case of evidence-free modelling is \citet{lee-etal-2020-language}, which relies on the information stored in the weights of large pre-trained language models instead of evidence documents. 

% Main challenges: Scale of retrievers, where to find data + constructed vs real, multimodality, source evaluation, mixed labels, misleading (but not false) claims, conveying findings

\subsection{Justification Production}
\label{section:modeling_justification_generation}

% Attention-based explanations
%As discussed in Section~\ref{section:definitions_justifications}, justification production inherits three large challenges from the literature on explainable machine learning -- readability, plausibility, and faithfulness. 
Approaches for justification production can be separated into three categories, which we examine along the three dimensions discussed in Section~\ref{section:definitions_justifications} --  readability, plausibility, and faithfulness. First, some models include components that can be analysed as justifications by human experts, primarily attention modules. \citet{Popat2018DeClarEDF} selected evidence tokens that have higher attention weights as explanations. Similarly, co-attention~\citep{Shu2019dEFENDEF, Lu2020GCANGC} and self-attention~\citep{Yang2019XFakeEF} were used to highlight the salient excerpts from the evidence. \citet{Wu2020DTCADT} further combined decision trees and attention weights to explain which tokens were salient, and how they influenced predictions. Recent studies have shown the use of attention as explanation to be problematic. Some tokens with high attention scores can be removed without affecting predictions, while some tokens with low (non-zero) scores turn out to be crucial~\citep{Jain2019AttentionIN, Serrano2019IsAI, Pruthi2020LearningTD}. 
%However, recent studies have argued that attention weights are only noisy predictors of intermediate components’ importance~\citep{Jain2019AttentionIN, Serrano2019IsAI, Pruthi2020LearningTD}. 
Explanations provided by attention may therefore not be sufficiently faithful. Furthermore, as they are difficult for non-experts and/or those not well-versed in the architecture of the model to grasp, they lack readability.

% Inherently explainable models
Another approach is to construct decision-making processes that can be fully grasped by human experts. Rule-based methods use Horn rules and knowledge bases to mine explanations~\citep{GadElrab2019ExFaKTAF,AhmadiLPS19}, which can be directly understood and verified. These rules are mined from a pre-constructed knowledge base, such as DBpedia~\citep{AuerBKLCI07}. This limits what can be fact-checked to claims which are representable as triples, and to information present in the (often manually curated) knowledge base.

% Generative explainability
Finally, some recent work has focused on building models which -- like human experts -- can generate %convincing
textual explanations for their decisions. %Generation-based methods formulate the explanation task as a text summarization problem. 
\citet{Atanasova2020GeneratingFC} used an extractive approach to generate summaries, 
% (using fact-checking articles as a form of ground-truth evidence), 
while \citet{kotonya2020explainable} adopted the abstractive approach. %Natural language explanations have a very high degree of readability, with textual justifications also being the primary method for journalists to communicate their fact-checks to the general public. 
A potential issue is that such models can generate explanations that do not represent their actual veracity prediction process, but which are nevertheless plausible with respect to the decision. This is especially an issue with abstractive models, where hallucinations can produce very misleading justifications~\citep{maynez-etal-2020-faithfulness}. Also, the model of \citet{Atanasova2020GeneratingFC} assumes fact-checking articles provided as input during inference, which is unrealistic.

%These models, as discussed in Section~\ref{section:definitions_justifications}, provide explainability beyond simply showing what evidence was used to reach a verdict -- they account explicitly, through different approaches, for the reasoning employed to reach the verdict.

% Main challenges: Where to get data, faithfulness, readability, black-box + large models

\section{Related Tasks}

\paragraph{Misinformation and Disinformation}
Misinformation is defined as constituting a claim that contradicts or distorts common understandings of verifiable facts~\citep{Guess2020MisinformationDA}. On the other hand, disinformation is defined as the subset of misinformation that is deliberately propagated. This is a question of intent: disinformation is meant to deceive, while misinformation may be inadvertent or unintentional~\citep{Tucker2018SocialMP}. Fact-checking can help detect misinformation, but not distinguish it from disinformation. A recent survey~\citep{alam2021survey} proposed to integrate both factuality and harmfulness into a framework for multi-modal disinformation detection. Although misinformation and conspiracy theories overlap conceptually, conspiracy theories do not hinge exclusively on the truth value of the claims being made, as they are sometimes proved to be true~\citep{sunstein2009conspiracy}. 
A related problem is \textit{propaganda detection}, which overlaps with disinformation detection, but also includes identifying particular techniques such as appeals to emotion, logical fallacies, whataboutery, or cherry-picking~\citep{Martino2020ASO}.

Propaganda and the deliberate or accidental dissemination of misleading information has been studied extensively. \citet{jowett2019propaganda} address the subject from a communications perspective, \citet{taylor2003munitions} provides a historical approach, and \citet{goldman2021epistemology} tackle the related subject of epistemology and trust in social settings from a philosophical perspective. For fact-checking and the identification of misinformation by journalists, we direct the reader to \citet{silverman2014verification} and \citet{borel2016chicago}.

\paragraph{Detecting Previously Fact-checked Claims}
While in this survey we focus on methods for verifying claims by finding the evidence rather than relying on previously conducted fact checks, misleading claims are often repeated~\citep{HassanZACJGHJKN17}; thus it is useful to detect whether a claim has already been fact-checked. \citet{shaar-etal-2020-known} formulated this task recently by as ranking, and constructed two datasets. The social media version of the task then featured at the shared task CheckThat!~\citep{BarrnCedeo2020OverviewOC,NakovMEBMSAHHBN21}. This task was also explored by~\citet{Vo2020WhereAT} from a multi-modal perspective, where %social media
claims about images were matched against previously fact-checked claims. More recently, \citet{sheng-etal-2021-article} and \citet{kazemi2021claim} constructed datasets for this task in languages beyond English. % (e.g\ Chinese, Hindi, etc).
%for detecting fact-checked claims beyond English. 
\citet{Hossain2020COVIDLiesDC} detected misinformation by adopting a similar strategy. If a tweet was matched to any known COVID-19 related misconceptions, then it would be classified as misinformative. Matching claims against previously verified ones is a simpler task that can often be reduced to sentence-level similarity~\citep{shaar-etal-2020-known}, which is well studied in the context of textual entailment. Nevertheless, new claims and evidence emerge regularly. Previous fact-checks can be useful, but they can become outdated and potentially misleading over time.

% \subsection{Fake News Detection}
% A term that has become strongly associated with fact-checking is ``fake news'', especially since its use in the context of the 2016 US presidential elections. However, it is used to label claims on various aspects not necessarily related to veracity as in fact-checking~\citep{Vosoughi2018TheSO}. The most prominent example of such usage of the term of fake news is its application to media organizations of opposing political sides. Therefore we avoid using this term in this survey.

\section{Research Challenges}
\label{sec:challenges}

\paragraph{Choice of Labels}

The use of fine-grained %truthfulness 
labels by %journalistic 
fact-checking organisations has recently come under criticism~\citep{uscinski2013epistemology}. In-between labels like \textit{``mostly true''} often represent ``meta-ratings'' for composite claims consisting of multiple elementary claims of different veracity. %, and rely on highly subjective judgments about how to weigh individual components,  
%in the final rating,
%and result in noise in the datasets used in system development. 
For example, a politician might claim improvements to unemployment and productivity% during their tenure in office
; if one part is true and the other false%the claim about unemployment is true and the claim about productivity is false
, a fact-checker might label the full statement \textit{``half true''}. % (or \textit{``half false''}). 
%When the decisions of journalists are used as ground-truths for automated fact checking systems, models inherit this issue.
Noisy labels resulting from composite claims could be avoided by intervening at the dataset creation stage to manually split such claims% into their parts
, or by learning to do so automatically. 
%However, as the use of such labels by journalists reveals, there is a clear demand for verification of complex, composite claims. Learning to automatically decompose and verify such claims would therefore ultimately be necessary.
The separation of claims into \textit{truth} and \textit{falsehood} can be too simplistic% to fully address real-world misinformation
, as true claims can still mislead% and deceive
. Examples include cherry-picking, where evidence is chosen to suggest a misleading \textit{trend}~\citep{asudeh2020cherrypicking}, and technical truth, where true information is presented in a way that misleads (e.g.\ \textit{``I have never lost a game of chess''} is also true if the speaker has never played chess). A major challenge %for automated fact-checkers 
is integrating analysis of such claims into the existing frameworks. This could involve %the introduction of 
new labels identifying specific forms of deception, as is done in propaganda detection~\citep{da-san-martino-etal-2020-semeval}, or a greater focus on producing justifications to show \textit{why} claims are misleading~\citep{Atanasova2020GeneratingFC, kotonya2020explainable}.

\paragraph{Sources \& Subjectivity}

%As we have discussed, fact-checking systems assume access to a trustworthy body of information for use as evidence. Realistically, the creation and curation of such a body is difficult -- 

Not all information is equally trustworthy, and sometimes trustworthy sources contradict each other. This challenges the assumptions made by most current fact-checking research relying on a single source considered authoritative, such as Wikipedia. 
%This presents several challenges for future research. 
Methods must be developed to address the presence of disagreeing or untrustworthy evidence. % cases where some untrustworthy information can appear in the evidence, and to handle disagreement between evidence documents.% in a way that is satisfactory to users. %, and to address the reality that different users will place different degrees of trust in many information sources. 
Recent work proposed integrating credibility assessment as a part of the fact-checking task~\citep{wu2020evidenceaware}. This could be done for example by assessing the agreement between evidence sources, or by assessing the degree to which sources 
%have in the past cohered
cohere with known facts~\citep{li2016survey, dong2015knowledgebased, zhang-etal-2019-evidence}. 
Similarly, 
%as we mentioned in Section~\ref{section:definitions_claim_detection}, 
check-worthiness is a subjective concept varying along axes including target audience, recency, and geography. One solution is to focus solely on objective checkability~\citep{konstantinovskiy2018towards}. However, the practical limitations of fact-checking (e.g.\ the deadlines of journalists and the time-constraints of media consumers) often force the use of a triage system%, where claims deemed more check-worthy are given primary importance
~\citep{borel2016chicago}. This can introduce biases regardless of the intentions of journalists and system-developers to use objective criteria~\citep{uscinski2013epistemology, uscinski2015epistemology}. Addressing this challenge will require the development of systems allowing for real-time interaction with users to take into account their evolving needs.
%user groups to update system choices based on the changing needs and trends of the information environment.

\paragraph{Dataset Artefacts \& Biases}
Synthetic datasets constructed through crowd-sourcing are common% in many NLP datasets, which are often constructed using crowd-sourcing
~\citep{zeichner2012,HermannKGEKSB15,Williams2018ABC}. It has been shown that models tend to rely on biases in these datasets, without learning the underlying task~\citep{gururangan-etal-2018-annotation, poliak-etal-2018-hypothesis, McCoy2019RightFT}. For fact-checking, \citet{Schuster2019TowardsDF} showed that the predictions of models trained on FEVER~\citep{Thorne2018FEVERAL} were largely driven by indicative claim words. The FEVER 2.0 shared task explored how to generate adversarial claims and build systems resilient to such attacks~\citep{Thorne2019TheFS}. Alleviating such biases and increasing the robustness to adversarial examples remains an open question. Potential solutions include leveraging better modelling approaches~\citep{Utama2020MindTT,Utama2020TowardsDN, karimi-mahabadi-etal-2020-end,Thorne2021ElasticWC}, collecting data by adversarial games~\citep{foolme2021}, or context-sensitive inference~\citep{vitaminc2021}.

\paragraph{Multimodality}
Information (either in claims or evidence) can be conveyed through multiple modalities such as text, tables, images, audio, or video. % e.g.\ claims based on doctored images or misleading graphs. 
%Furthermore, systems may be able to harvest additional data about sources, such as user profiles or social network topology. 
Though the majority of existing works have focused on text%ual misinformation
, some efforts also investigated how to incorporate multimodal information, including claim detection with misleading images~\citep{Zhang2018FauxBusterAC}, propaganda detection over mixed images and text~\citep{dimitrov-etal-2021-detecting}, and claim verification for images~\citep{Zlatkova2019FactCheckingMF,NakamuraLW20}. 
\citet{monti2019fake} argued that rumours should be seen as signals propagating through a social network. Rumour detection is therefore inherently multimodal, requiring analysis of both graph structure and text. Available multimodal corpora are either small in size~\citep{Zhang2018FauxBusterAC, Zlatkova2019FactCheckingMF} or constructed based on distant supervision% rather than human annotation
~\citep{NakamuraLW20}. The construction of large-scale annotated datasets paired with evidence beyond metadata will facilitate the development of multimodal fact-checking systems.

\paragraph{Multilinguality}
Claims can occur in multiple languages, often different from the one(s) evidence is available in, calling for multilingual fact-checking systems. While misinformation spans both geographic and linguistic boundaries, most work in the field has focused on English. A possible approach for multilingual verification is to use translation systems for existing methods~\citep{Dementieva2020FakeND}, but relevant datasets in more languages are necessary for testing multilingual models’ performance within each language, and ideally also for training. Currently, there exist a handful of datasets for factual verification in languages other than English~\citep{Baly2018IntegratingSD,Lillie2019JointRS, Khouja2020StancePA,shahifakecovid,NorregaardD21}, but they do not offer a cross-lingual setup. More recently, \citet{gupta-srikumar-2021-x} introduced a multilingual dataset covering 25 languages% (13 for training and 12 for testing)
, but found that adding training data from other languages did not improve performance. How to effectively align, coordinate, and leverage resources from different languages remains an open question. %Therefore, how to effectively align and coordinate the verification resources, and leverage them over different languages to improve fact-checking remains an open question. 
One promising direction is to distill knowledge from high-resource to low-resource languages~\citep{kazemi2021claim}.

\paragraph{Faithfulness}
A significant unaddressed challenge in justification production is faithfulness. % -- that is, the degree to which justifications match the reasoning employed by the analysed models.
%~\citep{jacovi2020towards}. For example, justifications produced using attention weights may ignore salient evidence~\citep{Jain2019AttentionIN}, while abstractively generated justifications may include hallucinations~\citep{maynez-etal-2020-faithfulness}. Faithfulness is difficult to evaluate for, as human evaluators and human-produced gold standards often struggle to separate highly plausible, unfaithful explanations from faithful ones~\citep{jacovi2020towards}. %That is, they struggle to identify convincing, reasonable justifications that nevertheless do not match the \textit{actual} behaviour of the model.
%
As we discuss in Section~\ref{section:modeling_justification_generation}, some justifications -- such as those generated abstractively~\citep{maynez-etal-2020-faithfulness} -- may not be faithful. This 
%A lack of faithfulness 
can be highly problematic, especially if these justifications are used to convince users of the validity of model predictions~\citep{lertvittayakumjorn-toni-2019-human}. Faithfulness is difficult to evaluate for, as human evaluators and human-produced gold standards often struggle to separate highly plausible, unfaithful explanations from faithful ones~\citep{jacovi2020towards}. %For example, a model suffering from hallucinations might produce more convincing hallucinations for predictions about a certain topic, making it harder to identify faulty model reasoning about that topic and thus introducing bias. 
In the model interpretability domain, several recent papers have introduced strategies for testing or guaranteeing faithfulness. These include introducing formal criteria which models should uphold~\citep{yu2019rethinking}, measuring the accuracy of predictions after removing some or all of the predicted non-salient input elements~\citep{yeh2019infidelity, deyoung-etal-2020-eraser, atanasova-etal-2020-diagnostic}, or disproving the faithfulness of techniques by counterexample~\citep{Jain2019AttentionIN, wiegreffe2019attention}. 
Further work is needed to develop such techniques for justification production.% fact verification, and to develop models with guarantees of faithfulness.

\paragraph{From Debunking to Early Intervention and Prebunking}
The prevailing application of automated fact-checking %verification
is to discover and intervene against circulating misinformation, also referred to as debunking. Efforts have been made to respond quickly after the appearance of a piece of misinformation~\citep{monti2019fake}, but common to all approaches is that intervention takes place \textit{reactively} after misinformation has already been introduced to the public.
NLP technology could also be leveraged in \textit{proactive} strategies. %There are several approaches which could be taken. 
Prior work has employed network analysis and similar techniques to identify key actors for intervention in social networks~\citep{farajtabar2017fake}; using NLP, such techniques could be extended to take into account the information shared by these actors, in addition to graph-based features~\citep{nakov2020can, Mu2020IdentifyingTU}. 
Another direction is to %identify potential misinformation before it can spread widely, and 
disseminate countermessaging before misinformation can spread widely%to the public before they encounter the misinformation ``in the wild''
; this is also known as \textit{pre}-bunking, and has been shown to be more effective than post-hoc debunking~\citep{vanderlinden2017inoculating,roozenbeek2020pre,lewandosky2021prebunking}. NLP could play a crucial role both in early detection and in the creation of relevant countermessaging. Finally, training people to \textit{create} misinformation has been shown to increase resistance towards false claims~\citep{roozenbeek2019game}. NLP could be used to facilitate this process, or to provide an adversarial opponent for gamifying the creation of misinformation. This could be seen as a form of dialogue agent to educate users, however there are as of yet no resources for the development of such systems.
%Current research in psychology has shown that it is much more effective to intervene preemptively by making the public aware of misinformation before they encounter it, i.e.\ \textit{pre}-bunking~\citep{vanderlinden2017inoculating,roozenbeek2020pre,lewandosky2021prebunking}. Training people to \textit{create} misinformation has also been shown to increase resistance towards false claims~\citep{roozenbeek2019game}. Leveraging NLP technology for prebunking is an unexplored and potentially very impactful direction for future research.

%\paragraph{Complex Claims}
%While the datasets we have discussed in this paper have yielded great results in terms of model development, many of them contain a known bias towards simpler claims~\citep{ostrowski2020multi}. Disputed real-world claims are often longer, require more reasoning steps (``hops'') to answer. Furthermore, as they often originate from spoken language or social media, they frequently contain semantically ambiguous statements that require context or world knowledge to resolve. Several recent papers have introduced models or datasets aimed towards addressing multi-hop fact verification~\citep{ostrowski2020multi, jiang2020hover}. Long or semantically complex claims have seen less attention, and the existing work in the multi-hop setting is preliminary.

\section{Conclusion}
\label{sec:conclusion}

We have reviewed and evaluated current automated fact-checking research by unifying the task formulations and methodologies across different research efforts into one framework comprising claim detection, evidence retrieval, verdict prediction, and justification production. Based on the proposed framework, we have provided an extensive overview of the existing datasets and modelling strategies. Finally, we have identified vital challenges for future research to address.

\section*{Acknowledgements}

Zhijiang Guo, Michael Schlichtkrull and Andreas Vlachos are supported by the ERC grant AVeriTeC (GA 865958), The latter is further supported by the EU H2020 grant MONITIO (GA 965576). The authors would like to thank Rami Aly, Christos Christodoulopoulos, Nedjma Ousidhoum, and James Thorne for useful comments and suggestions.

\bibliography{tacl2018}

\begin{thebibliography}{205}
\expandafter\ifx\csname natexlab\endcsname\relax\def\natexlab#1{#1}\fi

\bibitem[{Adair et~al.(2017)Adair, Li, Yang, and Yu}]{Adair2017ProgressT}
Bill Adair, Chengkai Li, Jun Yang, and Cong Yu. 2017.
\newblock Progress toward “the holy grail”: The continued quest to automate
  fact-checking.
\newblock In \emph{Proceedings of the 2017 Computation+Journalism Symposium}.

\bibitem[{Ahmadi et~al.(2019)Ahmadi, Lee, Papotti, and Saeed}]{AhmadiLPS19}
Naser Ahmadi, Joohyung Lee, Paolo Papotti, and Mohammed Saeed. 2019.
\newblock \href
  {https://truthandtrustonline.com/wp-content/uploads/2019/09/paper\_15.pdf}
  {Explainable fact checking with probabilistic answer set programming}.
\newblock In \emph{Proceedings of the 2019 Truth and Trust Online Conference
  {(TTO} 2019), London, UK, October 4-5, 2019}.

\bibitem[{Aker et~al.(2017)Aker, Derczynski, and Bontcheva}]{Aker2017SimpleOS}
Ahmet Aker, Leon Derczynski, and Kalina Bontcheva. 2017.
\newblock \href {https://doi.org/10.26615/978-954-452-049-6_005} {Simple open
  stance classification for rumour analysis}.
\newblock In \emph{Proceedings of the International Conference Recent Advances
  in Natural Language Processing, {RANLP} 2017}, pages 31--39, Varna, Bulgaria.
  INCOMA Ltd.

\bibitem[{Alam et~al.(2021)Alam, Cresci, Chakraborty, Silvestri, Dimitrov,
  Martino, Shaar, Firooz, and Nakov}]{alam2021survey}
Firoj Alam, Stefano Cresci, Tanmoy Chakraborty, Fabrizio Silvestri, Dimiter
  Dimitrov, Giovanni Da~San Martino, Shaden Shaar, Hamed Firooz, and Preslav
  Nakov. 2021.
\newblock A survey on multimodal disinformation detection.
\newblock \emph{arXiv preprint arXiv:2103.12541}.

\bibitem[{Alhindi et~al.(2018)Alhindi, Petridis, and
  Muresan}]{Alhindi2018WhereIY}
Tariq Alhindi, Savvas Petridis, and Smaranda Muresan. 2018.
\newblock \href {https://doi.org/10.18653/v1/W18-5513} {Where is your evidence:
  Improving fact-checking by justification modeling}.
\newblock In \emph{Proceedings of the First Workshop on Fact Extraction and
  {VER}ification ({FEVER})}, pages 85--90, Brussels, Belgium. Association for
  Computational Linguistics.

\bibitem[{Allein et~al.(2021)Allein, Augenstein, and Moens}]{allein2021time}
Liesbeth Allein, Isabelle Augenstein, and Marie-Francine Moens. 2021.
\newblock {Time-Aware Evidence Ranking for Fact-Checking}.
\newblock \emph{Web Semantics}.

\bibitem[{Aly et~al.(2021)Aly, Guo, Schlichtkrull, Thorne, Vlachos,
  Christodoulopoulos, Cocarascu, and Mittal}]{Aly2021FEVEROUSFE}
Rami Aly, Zhijiang Guo, M.~Schlichtkrull, James Thorne, Andreas Vlachos,
  Christos Christodoulopoulos, O.~Cocarascu, and Arpit Mittal. 2021.
\newblock {FEVEROUS}: {Fact Extraction} and {VERification} over unstructured
  and structured information.
\newblock \emph{35th Conference on Neural Information Processing Systems
  (NeurIPS 2021) Track on Datasets and Benchmarks}.

\bibitem[{Asudeh et~al.(2020)Asudeh, Jagadish, Wu, and
  Yu}]{asudeh2020cherrypicking}
Abolfazl Asudeh, H.~V. Jagadish, You~(Will) Wu, and Cong Yu. 2020.
\newblock \href {https://doi.org/10.14778/3380750.3380762} {On detecting
  cherry-picked trendlines}.
\newblock \emph{Proceedings of the VLDB Endowment}, 13(6):939–952.

\bibitem[{Atanasova et~al.(2018)Atanasova, M{\`{a}}rquez,
  Barr{\'{o}}n{-}Cede{\~{n}}o, Elsayed, Suwaileh, Zaghouani, Kyuchukov,
  Martino, and Nakov}]{Atanasova2018OverviewOT}
Pepa Atanasova, Llu{\'{\i}}s M{\`{a}}rquez, Alberto
  Barr{\'{o}}n{-}Cede{\~{n}}o, Tamer Elsayed, Reem Suwaileh, Wajdi Zaghouani,
  Spas Kyuchukov, Giovanni Da~San Martino, and Preslav Nakov. 2018.
\newblock \href {http://ceur-ws.org/Vol-2125/invited\_paper\_13.pdf} {Overview
  of the {CLEF-2018} {CheckThat!} lab on automatic identification and
  verification of political claims. task 1: Check-worthiness}.
\newblock In \emph{Working Notes of {CLEF} 2018 - Conference and Labs of the
  Evaluation Forum, Avignon, France, September 10-14, 2018}, volume 2125 of
  \emph{{CEUR} Workshop Proceedings}. CEUR-WS.org.

\bibitem[{Atanasova et~al.(2020{\natexlab{a}})Atanasova, Simonsen, Lioma, and
  Augenstein}]{atanasova-etal-2020-diagnostic}
Pepa Atanasova, Jakob~Grue Simonsen, Christina Lioma, and Isabelle Augenstein.
  2020{\natexlab{a}}.
\newblock \href {https://doi.org/10.18653/v1/2020.emnlp-main.263} {A diagnostic
  study of explainability techniques for text classification}.
\newblock In \emph{Proceedings of the 2020 Conference on Empirical Methods in
  Natural Language Processing (EMNLP)}, pages 3256--3274, Online. Association
  for Computational Linguistics.

\bibitem[{Atanasova et~al.(2020{\natexlab{b}})Atanasova, Simonsen, Lioma, and
  Augenstein}]{Atanasova2020GeneratingFC}
Pepa Atanasova, Jakob~Grue Simonsen, Christina Lioma, and Isabelle Augenstein.
  2020{\natexlab{b}}.
\newblock \href {https://doi.org/10.18653/v1/2020.acl-main.656} {Generating
  fact checking explanations}.
\newblock In \emph{Proceedings of the 58th Annual Meeting of the Association
  for Computational Linguistics}, pages 7352--7364, Online. Association for
  Computational Linguistics.

\bibitem[{Auer et~al.(2007)Auer, Bizer, Kobilarov, Lehmann, Cyganiak, and
  Ives}]{AuerBKLCI07}
S{\"{o}}ren Auer, Christian Bizer, Georgi Kobilarov, Jens Lehmann, Richard
  Cyganiak, and Zachary~G. Ives. 2007.
\newblock \href {https://doi.org/10.1007/978-3-540-76298-0\_52} {{DBpedia}: {A}
  nucleus for a web of open data}.
\newblock In \emph{The Semantic Web, 6th International Semantic Web Conference,
  2nd Asian Semantic Web Conference, {ISWC} 2007 + {ASWC} 2007, Busan, Korea,
  November 11-15, 2007}, volume 4825 of \emph{Lecture Notes in Computer
  Science}, pages 722--735. Springer.

\bibitem[{Augenstein et~al.(2019)Augenstein, Lioma, Wang, Chaves~Lima, Hansen,
  Hansen, and Simonsen}]{Augenstein2019MultiFCAR}
Isabelle Augenstein, Christina Lioma, Dongsheng Wang, Lucas Chaves~Lima, Casper
  Hansen, Christian Hansen, and Jakob~Grue Simonsen. 2019.
\newblock \href {https://doi.org/10.18653/v1/D19-1475} {{M}ulti{FC}: A
  real-world multi-domain dataset for evidence-based fact checking of claims}.
\newblock In \emph{Proceedings of the 2019 Conference on Empirical Methods in
  Natural Language Processing and the 9th International Joint Conference on
  Natural Language Processing (EMNLP-IJCNLP)}, pages 4685--4697, Hong Kong,
  China. Association for Computational Linguistics.

\bibitem[{Bachenko et~al.(2008)Bachenko, Fitzpatrick, and
  Schonwetter}]{bachenko-etal-2008-verification}
Joan Bachenko, Eileen Fitzpatrick, and Michael Schonwetter. 2008.
\newblock \href {https://www.aclweb.org/anthology/C08-1006} {Verification and
  implementation of language-based deception indicators in civil and criminal
  narratives}.
\newblock In \emph{Proceedings of the 22nd International Conference on
  Computational Linguistics (Coling 2008)}, pages 41--48, Manchester, UK.
  Coling 2008 Organizing Committee.

\bibitem[{Baly et~al.(2018)Baly, Mohtarami, Glass, M{\`a}rquez, Moschitti, and
  Nakov}]{Baly2018IntegratingSD}
Ramy Baly, Mitra Mohtarami, James Glass, Llu{\'\i}s M{\`a}rquez, Alessandro
  Moschitti, and Preslav Nakov. 2018.
\newblock \href {https://doi.org/10.18653/v1/N18-2004} {Integrating stance
  detection and fact checking in a unified corpus}.
\newblock In \emph{Proceedings of the 2018 Conference of the North {A}merican
  Chapter of the Association for Computational Linguistics: Human Language
  Technologies, Volume 2 (Short Papers)}, pages 21--27, New Orleans, Louisiana.
  Association for Computational Linguistics.

\bibitem[{Barnoy and Reich(2019)}]{Barnoy2019}
Aviv Barnoy and Zvi Reich. 2019.
\newblock \href {https://doi.org/10.1080/1461670X.2019.1593881} {{The When,
  Why, How and So-What of Verifications}}.
\newblock \emph{Journalism Studies}, 20(16):2312--2330.

\bibitem[{Barr{\'{o}}n{-}Cede{\~{n}}o et~al.(2020)Barr{\'{o}}n{-}Cede{\~{n}}o,
  Elsayed, Nakov, Martino, Hasanain, Suwaileh, Haouari, Babulkov, Hamdan,
  Nikolov, Shaar, and Ali}]{BarrnCedeo2020OverviewOC}
Alberto Barr{\'{o}}n{-}Cede{\~{n}}o, Tamer Elsayed, Preslav Nakov, Giovanni
  Da~San Martino, Maram Hasanain, Reem Suwaileh, Fatima Haouari, Nikolay
  Babulkov, Bayan Hamdan, Alex Nikolov, Shaden Shaar, and Zien~Sheikh Ali.
  2020.
\newblock \href {https://doi.org/10.1007/978-3-030-58219-7\_17} {Overview of
  {CheckThat!} 2020: Automatic identification and verification of claims in
  social media}.
\newblock In \emph{Experimental {IR} Meets Multilinguality, Multimodality, and
  Interaction - 11th International Conference of the {CLEF} Association, {CLEF}
  2020, Thessaloniki, Greece, September 22-25, 2020, Proceedings}, volume 12260
  of \emph{Lecture Notes in Computer Science}, pages 215--236. Springer.

\bibitem[{Barr{\'{o}}n{-}Cede{\~{n}}o et~al.(2018)Barr{\'{o}}n{-}Cede{\~{n}}o,
  Elsayed, Suwaileh, M{\`{a}}rquez, Atanasova, Zaghouani, Kyuchukov, Martino,
  and Nakov}]{Barron-CedenoES18}
Alberto Barr{\'{o}}n{-}Cede{\~{n}}o, Tamer Elsayed, Reem Suwaileh, Llu{\'{\i}}s
  M{\`{a}}rquez, Pepa Atanasova, Wajdi Zaghouani, Spas Kyuchukov, Giovanni
  Da~San Martino, and Preslav Nakov. 2018.
\newblock \href {http://ceur-ws.org/Vol-2125/invited\_paper\_14.pdf} {Overview
  of the {CLEF-2018} {CheckThat!} lab on automatic identification and
  verification of political claims. task 2: Factuality}.
\newblock In \emph{Working Notes of {CLEF} 2018 - Conference and Labs of the
  Evaluation Forum, Avignon, France, September 10-14, 2018}, volume 2125 of
  \emph{{CEUR} Workshop Proceedings}. CEUR-WS.org.

\bibitem[{Bollacker et~al.(2008)Bollacker, Evans, Paritosh, Sturge, and
  Taylor}]{BollackerEPST08}
Kurt~D. Bollacker, Colin Evans, Praveen Paritosh, Tim Sturge, and Jamie Taylor.
  2008.
\newblock \href {https://doi.org/10.1145/1376616.1376746} {Freebase: a
  collaboratively created graph database for structuring human knowledge}.
\newblock In \emph{Proceedings of the {ACM} {SIGMOD} International Conference
  on Management of Data, {SIGMOD} 2008, Vancouver, BC, Canada, June 10-12,
  2008}, pages 1247--1250. {ACM}.

\bibitem[{Bordes et~al.(2013)Bordes, Usunier, Garc{\'{\i}}a{-}Dur{\'{a}}n,
  Weston, and Yakhnenko}]{bordes2013translating}
Antoine Bordes, Nicolas Usunier, Alberto Garc{\'{\i}}a{-}Dur{\'{a}}n, Jason
  Weston, and Oksana Yakhnenko. 2013.
\newblock \href
  {https://proceedings.neurips.cc/paper/2013/hash/1cecc7a77928ca8133fa24680a88d2f9-Abstract.html}
  {Translating embeddings for modeling multi-relational data}.
\newblock In \emph{Advances in Neural Information Processing Systems 26: 27th
  Annual Conference on Neural Information Processing Systems 2013. Proceedings
  of a meeting held December 5-8, 2013, Lake Tahoe, Nevada, United States},
  pages 2787--2795.

\bibitem[{Borel(2016)}]{borel2016chicago}
Brooke Borel. 2016.
\newblock \emph{The Chicago Guide to Fact-checking}.
\newblock University of Chicago Press.

\bibitem[{Bowman et~al.(2015)Bowman, Angeli, Potts, and
  Manning}]{Bowman2015ALA}
Samuel~R. Bowman, Gabor Angeli, Christopher Potts, and Christopher~D. Manning.
  2015.
\newblock \href {https://doi.org/10.18653/v1/D15-1075} {A large annotated
  corpus for learning natural language inference}.
\newblock In \emph{Proceedings of the 2015 Conference on Empirical Methods in
  Natural Language Processing}, pages 632--642, Lisbon, Portugal. Association
  for Computational Linguistics.

\bibitem[{Brown et~al.(2020)Brown, Mann, Ryder, Subbiah, Kaplan, Dhariwal,
  Neelakantan, Shyam, Sastry, Askell, Agarwal, Herbert{-}Voss, Krueger,
  Henighan, Child, Ramesh, Ziegler, Wu, Winter, Hesse, Chen, Sigler, Litwin,
  Gray, Chess, Clark, Berner, McCandlish, Radford, Sutskever, and
  Amodei}]{Brown2020LanguageMA}
Tom~B. Brown, Benjamin Mann, Nick Ryder, Melanie Subbiah, Jared Kaplan,
  Prafulla Dhariwal, Arvind Neelakantan, Pranav Shyam, Girish Sastry, Amanda
  Askell, Sandhini Agarwal, Ariel Herbert{-}Voss, Gretchen Krueger, Tom
  Henighan, Rewon Child, Aditya Ramesh, Daniel~M. Ziegler, Jeffrey Wu, Clemens
  Winter, Christopher Hesse, Mark Chen, Eric Sigler, Mateusz Litwin, Scott
  Gray, Benjamin Chess, Jack Clark, Christopher Berner, Sam McCandlish, Alec
  Radford, Ilya Sutskever, and Dario Amodei. 2020.
\newblock \href
  {https://proceedings.neurips.cc/paper/2020/hash/1457c0d6bfcb4967418bfb8ac142f64a-Abstract.html}
  {Language models are few-shot learners}.
\newblock In \emph{Advances in Neural Information Processing Systems 33: Annual
  Conference on Neural Information Processing Systems 2020, NeurIPS 2020,
  December 6-12, 2020, virtual}.

\bibitem[{Buntain and Golbeck(2017)}]{code2017automatically}
Cody Buntain and Jennifer Golbeck. 2017.
\newblock \href {https://doi.org/10.1109/SmartCloud.2017.40} {Automatically
  identifying fake news in popular twitter threads}.
\newblock In \emph{2017 IEEE International Conference on Smart Cloud
  (SmartCloud)}, pages 208--215. IEEE.

\bibitem[{Chen et~al.(2020)Chen, Wang, Chen, Zhang, Wang, Li, Zhou, and
  Wang}]{Chen2020TabFactAL}
Wenhu Chen, Hongmin Wang, Jianshu Chen, Yunkai Zhang, Hong Wang, Shiyang Li,
  Xiyou Zhou, and William~Yang Wang. 2020.
\newblock \href {https://openreview.net/forum?id=rkeJRhNYDH} {{TabFact}: {A}
  large-scale dataset for table-based fact verification}.
\newblock In \emph{8th International Conference on Learning Representations,
  {ICLR} 2020}, Addis Ababa, Ethiopia.

\bibitem[{Ciampaglia et~al.(2015)Ciampaglia, Shiralkar, Rocha, Bollen, Menczer,
  and Flammini}]{Ciampaglia2015ComputationalFC}
Giovanni~Luca Ciampaglia, Prashant Shiralkar, Luis~M Rocha, Johan Bollen,
  Filippo Menczer, and Alessandro Flammini. 2015.
\newblock Computational fact checking from knowledge networks.
\newblock \emph{PloS one}, 10(6):e0128193.

\bibitem[{Cohen et~al.(2011)Cohen, Li, Yang, and Yu}]{Cohen2011ComputationalJA}
Sarah Cohen, Chengkai Li, Jun Yang, and Cong Yu. 2011.
\newblock \href {http://cidrdb.org/cidr2011/Papers/CIDR11\_Paper17.pdf}
  {Computational journalism: {A} call to arms to database researchers}.
\newblock In \emph{{CIDR} 2011, Fifth Biennial Conference on Innovative Data
  Systems Research, Asilomar, CA, USA, January 9-12, 2011, Online Proceedings},
  pages 148--151. www.cidrdb.org.

\bibitem[{Da~San~Martino et~al.(2020{\natexlab{a}})Da~San~Martino,
  Barr{\'o}n-Cede{\~n}o, Wachsmuth, Petrov, and
  Nakov}]{da-san-martino-etal-2020-semeval}
Giovanni Da~San~Martino, Alberto Barr{\'o}n-Cede{\~n}o, Henning Wachsmuth,
  Rostislav Petrov, and Preslav Nakov. 2020{\natexlab{a}}.
\newblock \href {https://www.aclweb.org/anthology/2020.semeval-1.186}
  {{S}em{E}val-2020 task 11: Detection of propaganda techniques in news
  articles}.
\newblock In \emph{Proceedings of the Fourteenth Workshop on Semantic
  Evaluation}, pages 1377--1414, Barcelona (online). International Committee
  for Computational Linguistics.

\bibitem[{Da~San~Martino et~al.(2020{\natexlab{b}})Da~San~Martino, Cresci,
  Barr{\'{o}}n{-}Cede{\~{n}}o, Yu, Di~Pietro, and Nakov}]{Martino2020ASO}
Giovanni Da~San~Martino, Stefano Cresci, Alberto Barr{\'{o}}n{-}Cede{\~{n}}o,
  Seunghak Yu, Roberto Di~Pietro, and Preslav Nakov. 2020{\natexlab{b}}.
\newblock \href {https://doi.org/10.24963/ijcai.2020/672} {A survey on
  computational propaganda detection}.
\newblock In \emph{Proceedings of the Twenty-Ninth International Joint
  Conference on Artificial Intelligence, {IJCAI} 2020}, pages 4826--4832.
  ijcai.org.

\bibitem[{Dagan et~al.(2010)Dagan, Dolan, Magnini, and Roth}]{DaganDMR10}
Ido Dagan, Bill Dolan, Bernardo Magnini, and Dan Roth. 2010.
\newblock \href {https://doi.org/10.1017/S1351324909990234} {Recognizing
  textual entailment: Rational, evaluation and approaches}.
\newblock \emph{Natural Language Engingeering}, 16(1):105.

\bibitem[{Dementieva and Panchenko(2020)}]{Dementieva2020FakeND}
Daryna Dementieva and A.~Panchenko. 2020.
\newblock Fake news detection using multilingual evidence.
\newblock \emph{2020 IEEE 7th International Conference on Data Science and
  Advanced Analytics (DSAA)}, pages 775--776.

\bibitem[{Derczynski et~al.(2017)Derczynski, Bontcheva, Liakata, Procter, Wong
  Sak~Hoi, and Zubiaga}]{Derczynski2017SemEval2017T8}
Leon Derczynski, Kalina Bontcheva, Maria Liakata, Rob Procter, Geraldine Wong
  Sak~Hoi, and Arkaitz Zubiaga. 2017.
\newblock \href {https://doi.org/10.18653/v1/S17-2006} {{S}em{E}val-2017 task
  8: {R}umour{E}val: Determining rumour veracity and support for rumours}.
\newblock In \emph{Proceedings of the 11th International Workshop on Semantic
  Evaluation ({S}em{E}val-2017)}, pages 69--76, Vancouver, Canada. Association
  for Computational Linguistics.

\bibitem[{Devlin et~al.(2019)Devlin, Chang, Lee, and
  Toutanova}]{Devlin2019BERTPO}
Jacob Devlin, Ming-Wei Chang, Kenton Lee, and Kristina Toutanova. 2019.
\newblock \href {https://doi.org/10.18653/v1/N19-1423} {{BERT}: Pre-training of
  deep bidirectional transformers for language understanding}.
\newblock In \emph{Proceedings of the 2019 Conference of the North {A}merican
  Chapter of the Association for Computational Linguistics: Human Language
  Technologies, Volume 1 (Long and Short Papers)}, pages 4171--4186,
  Minneapolis, Minnesota. Association for Computational Linguistics.

\bibitem[{DeYoung et~al.(2020)DeYoung, Jain, Rajani, Lehman, Xiong, Socher, and
  Wallace}]{deyoung-etal-2020-eraser}
Jay DeYoung, Sarthak Jain, Nazneen~Fatema Rajani, Eric Lehman, Caiming Xiong,
  Richard Socher, and Byron~C. Wallace. 2020.
\newblock \href {https://doi.org/10.18653/v1/2020.acl-main.408} {{ERASER}: {A}
  benchmark to evaluate rationalized {NLP} models}.
\newblock In \emph{Proceedings of the 58th Annual Meeting of the Association
  for Computational Linguistics}, pages 4443--4458, Online. Association for
  Computational Linguistics.

\bibitem[{Diggelmann et~al.(2020)Diggelmann, Boyd{-}Graber, Bulian, Ciaramita,
  and Leippold}]{diggelmann2020}
Thomas Diggelmann, Jordan~L. Boyd{-}Graber, Jannis Bulian, Massimiliano
  Ciaramita, and Markus Leippold. 2020.
\newblock \href {http://arxiv.org/abs/2012.00614} {{CLIMATE-FEVER:} {A} dataset
  for verification of real-world climate claims}.
\newblock \emph{CoRR}, abs/2012.00614.

\bibitem[{Dimitrov et~al.(2021)Dimitrov, Bin~Ali, Shaar, Alam, Silvestri,
  Firooz, Nakov, and Da~San~Martino}]{dimitrov-etal-2021-detecting}
Dimitar Dimitrov, Bishr Bin~Ali, Shaden Shaar, Firoj Alam, Fabrizio Silvestri,
  Hamed Firooz, Preslav Nakov, and Giovanni Da~San~Martino. 2021.
\newblock \href {https://doi.org/10.18653/v1/2021.acl-long.516} {Detecting
  propaganda techniques in memes}.
\newblock In \emph{Proceedings of the 59th Annual Meeting of the Association
  for Computational Linguistics and the 11th International Joint Conference on
  Natural Language Processing (Volume 1: Long Papers)}, pages 6603--6617,
  Online. Association for Computational Linguistics.

\bibitem[{Dong et~al.(2015)Dong, Gabrilovich, Murphy, Dang, Horn, Lugaresi,
  Sun, and Zhang}]{dong2015knowledgebased}
Xin~Luna Dong, Evgeniy Gabrilovich, Kevin Murphy, Van Dang, Wilko Horn, Camillo
  Lugaresi, Shaohua Sun, and Wei Zhang. 2015.
\newblock \href {https://doi.org/10.14778/2777598.2777603} {Knowledge-based
  trust: Estimating the trustworthiness of web sources}.
\newblock \emph{Proceedings of the {VLDB} Endowment}, 8(9):938–949.

\bibitem[{Dungs et~al.(2018)Dungs, Aker, Fuhr, and Bontcheva}]{Dungs2018CanRS}
Sebastian Dungs, Ahmet Aker, Norbert Fuhr, and Kalina Bontcheva. 2018.
\newblock \href {https://www.aclweb.org/anthology/C18-1284} {Can rumour stance
  alone predict veracity?}
\newblock In \emph{Proceedings of the 27th International Conference on
  Computational Linguistics}, pages 3360--3370, Santa Fe, New Mexico, USA.
  Association for Computational Linguistics.

\bibitem[{Eisenschlos et~al.(2021)Eisenschlos, Dhingra, Bulian,
  B{\"o}rschinger, and Boyd-Graber}]{foolme2021}
Julian Eisenschlos, Bhuwan Dhingra, Jannis Bulian, Benjamin B{\"o}rschinger,
  and Jordan Boyd-Graber. 2021.
\newblock \href {https://www.aclweb.org/anthology/2021.naacl-main.32} {{Fool Me
  Twice}: Entailment from {W}ikipedia gamification}.
\newblock In \emph{Proceedings of the 2021 Conference of the North American
  Chapter of the Association for Computational Linguistics: Human Language
  Technologies}, pages 352--365, Online. Association for Computational
  Linguistics.

\bibitem[{Eisenschlos et~al.(2020)Eisenschlos, Krichene, and
  M{\"u}ller}]{eisenschlos-etal-2020-understanding}
Julian Eisenschlos, Syrine Krichene, and Thomas M{\"u}ller. 2020.
\newblock \href {https://doi.org/10.18653/v1/2020.findings-emnlp.27}
  {Understanding tables with intermediate pre-training}.
\newblock In \emph{Findings of the Association for Computational Linguistics:
  EMNLP 2020}, pages 281--296, Online. Association for Computational
  Linguistics.

\bibitem[{Enayet and El-Beltagy(2017)}]{Enayet2017NileTMRGAS}
Omar Enayet and Samhaa~R. El-Beltagy. 2017.
\newblock \href {https://doi.org/10.18653/v1/S17-2082} {{N}ile{TMRG} at
  {S}em{E}val-2017 task 8: Determining rumour and veracity support for rumours
  on {T}witter.}
\newblock In \emph{Proceedings of the 11th International Workshop on Semantic
  Evaluation ({S}em{E}val-2017)}, pages 470--474, Vancouver, Canada.
  Association for Computational Linguistics.

\bibitem[{Fan et~al.(2020)Fan, Piktus, Petroni, Wenzek, Saeidi, Vlachos,
  Bordes, and Riedel}]{Fan2020GeneratingFC}
Angela Fan, Aleksandra Piktus, Fabio Petroni, Guillaume Wenzek, Marzieh Saeidi,
  Andreas Vlachos, Antoine Bordes, and Sebastian Riedel. 2020.
\newblock \href {https://doi.org/10.18653/v1/2020.emnlp-main.580} {Generating
  fact checking briefs}.
\newblock In \emph{Proceedings of the 2020 Conference on Empirical Methods in
  Natural Language Processing (EMNLP)}, pages 7147--7161, Online. Association
  for Computational Linguistics.

\bibitem[{Farajtabar et~al.(2017)Farajtabar, Yang, Ye, Xu, Trivedi, Khalil, Li,
  Song, and Zha}]{farajtabar2017fake}
Mehrdad Farajtabar, Jiachen Yang, Xiaojing Ye, Huan Xu, Rakshit Trivedi, Elias
  Khalil, Shuang Li, Le~Song, and Hongyuan Zha. 2017.
\newblock \href {http://proceedings.mlr.press/v70/farajtabar17a.html} {Fake
  news mitigation via point process based intervention}.
\newblock In \emph{Proceedings of the 34th International Conference on Machine
  Learning}, volume~70 of \emph{Proceedings of Machine Learning Research},
  pages 1097--1106. PMLR.

\bibitem[{Ferreira and Vlachos(2016)}]{Ferreira2016EmergentAN}
William Ferreira and Andreas Vlachos. 2016.
\newblock \href {https://doi.org/10.18653/v1/N16-1138} {{E}mergent: a novel
  data-set for stance classification}.
\newblock In \emph{Proceedings of the 2016 Conference of the North {A}merican
  Chapter of the Association for Computational Linguistics: Human Language
  Technologies}, pages 1163--1168, San Diego, California. Association for
  Computational Linguistics.

\bibitem[{Flew et~al.(2010)Flew, Spurgeon, Daniel, and Swift}]{Flew2010THEPO}
Terry Flew, Christina Spurgeon, Anna Daniel, and Adam Swift. 2010.
\newblock The promise of computational journalism.
\newblock \emph{Journalism Practice}, 6:157 -- 171.

\bibitem[{Gad{-}Elrab et~al.(2019)Gad{-}Elrab, Stepanova, Urbani, and
  Weikum}]{GadElrab2019ExFaKTAF}
Mohamed~H. Gad{-}Elrab, Daria Stepanova, Jacopo Urbani, and Gerhard Weikum.
  2019.
\newblock \href {https://doi.org/10.1145/3289600.3290996} {{ExFaKT}: {A}
  framework for explaining facts over knowledge graphs and text}.
\newblock In \emph{Proceedings of the Twelfth {ACM} International Conference on
  Web Search and Data Mining, {WSDM} 2019, Melbourne, VIC, Australia, February
  11-15, 2019}, pages 87--95. {ACM}.

\bibitem[{Gencheva et~al.(2017)Gencheva, Nakov, M{\`a}rquez,
  Barr{\'o}n-Cede{\~n}o, and Koychev}]{Gencheva2017ACA}
Pepa Gencheva, Preslav Nakov, Llu{\'\i}s M{\`a}rquez, Alberto
  Barr{\'o}n-Cede{\~n}o, and Ivan Koychev. 2017.
\newblock \href {https://doi.org/10.26615/978-954-452-049-6_037} {A
  context-aware approach for detecting worth-checking claims in political
  debates}.
\newblock In \emph{Proceedings of the International Conference Recent Advances
  in Natural Language Processing, {RANLP} 2017}, pages 267--276, Varna,
  Bulgaria. INCOMA Ltd.

\bibitem[{Godler and Reich(2017)}]{godler2017journalistic}
Yigal Godler and Zvi Reich. 2017.
\newblock Journalistic evidence: Cross-verification as a constituent of
  mediated knowledge.
\newblock \emph{Journalism}, 18(5):558--574.

\bibitem[{Goldman and O’Connor(2021)}]{goldman2021epistemology}
Alvin Goldman and Cailin O’Connor. 2021.
\newblock {Social Epistemology}.
\newblock In Edward~N. Zalta, editor, \emph{The {Stanford} Encyclopedia of
  Philosophy}, {S}pring 2021 edition. Metaphysics Research Lab, Stanford
  University.

\bibitem[{Gorrell et~al.(2019)Gorrell, Aker, Bontcheva, Derczynski, Kochkina,
  Liakata, and Zubiaga}]{GorrellABDKLZ19}
Genevieve Gorrell, Ahmet Aker, Kalina Bontcheva, Leon Derczynski, Elena
  Kochkina, Maria Liakata, and Arkaitz Zubiaga. 2019.
\newblock \href {https://doi.org/10.18653/v1/s19-2147} {{SemEval-2019} task 7:
  {RumourEval}, determining rumour veracity and support for rumours}.
\newblock In \emph{Proceedings of the 13th International Workshop on Semantic
  Evaluation, SemEval@NAACL-HLT 2019, Minneapolis, MN, USA, June 6-7, 2019},
  pages 845--854. Association for Computational Linguistics.

\bibitem[{Graves(2018)}]{2018graves}
Lucas Graves. 2018.
\newblock Understanding the promise and limits of automated fact-checking.
\newblock \emph{Reuters Institute for the Study of Journalism}.

\bibitem[{Gruppi et~al.(2020)Gruppi, Horne, and Adali}]{gruppi2020}
Maur{\'{\i}}cio Gruppi, Benjamin~D. Horne, and Sibel Adali. 2020.
\newblock \href {http://arxiv.org/abs/2003.08444} {{NELA-GT-2019:} {A} large
  multi-labelled news dataset for the study of misinformation in news
  articles}.
\newblock \emph{CoRR}, abs/2003.08444.

\bibitem[{Gruppi et~al.(2021)Gruppi, Horne, and Adali}]{gruppi2021}
Maur{\'{\i}}cio Gruppi, Benjamin~D. Horne, and Sibel Adali. 2021.
\newblock \href {http://arxiv.org/abs/2102.04567} {{NELA-GT-2020:} {A} large
  multi-labelled news dataset for the study of misinformation in news
  articles}.
\newblock \emph{CoRR}, abs/2102.04567.

\bibitem[{Guess and Lyons(2020)}]{Guess2020MisinformationDA}
Andrew~M. Guess and Benjamin~A. Lyons. 2020.
\newblock Misinformation, disinformation, and online propaganda.
\newblock In Nathaniel Persily and Joshua~A. Tucker, editors, \emph{Social
  media and democracy: the state of the field, prospects for reform}, pages
  10--33. Cambridge University Press.

\bibitem[{Guo et~al.(2018)Guo, Cao, Zhang, Guo, and Li}]{Guo2018RumorDW}
Han Guo, Juan Cao, Yazi Zhang, Junbo Guo, and Jintao Li. 2018.
\newblock \href {https://doi.org/10.1145/3269206.3271709} {Rumor detection with
  hierarchical social attention network}.
\newblock In \emph{Proceedings of the 27th {ACM} International Conference on
  Information and Knowledge Management, {CIKM} 2018, Torino, Italy, October
  22-26, 2018}, pages 943--951. {ACM}.

\bibitem[{Gupta and Srikumar(2021)}]{gupta-srikumar-2021-x}
Ashim Gupta and Vivek Srikumar. 2021.
\newblock \href {https://doi.org/10.18653/v1/2021.acl-short.86} {{X-Fact}: A
  new benchmark dataset for multilingual fact checking}.
\newblock In \emph{Proceedings of the 59th Annual Meeting of the Association
  for Computational Linguistics and the 11th International Joint Conference on
  Natural Language Processing (Volume 2: Short Papers)}, pages 675--682,
  Online. Association for Computational Linguistics.

\bibitem[{Gupta et~al.(2020)Gupta, Mehta, Nokhiz, and
  Srikumar}]{Gupta2020INFOTABSIO}
Vivek Gupta, Maitrey Mehta, Pegah Nokhiz, and Vivek Srikumar. 2020.
\newblock \href {https://doi.org/10.18653/v1/2020.acl-main.210} {{INFOTABS}:
  Inference on tables as semi-structured data}.
\newblock In \emph{Proceedings of the 58th Annual Meeting of the Association
  for Computational Linguistics}, pages 2309--2324, Online. Association for
  Computational Linguistics.

\bibitem[{Gururangan et~al.(2018)Gururangan, Swayamdipta, Levy, Schwartz,
  Bowman, and Smith}]{gururangan-etal-2018-annotation}
Suchin Gururangan, Swabha Swayamdipta, Omer Levy, Roy Schwartz, Samuel Bowman,
  and Noah~A. Smith. 2018.
\newblock \href {https://doi.org/10.18653/v1/N18-2017} {Annotation artifacts in
  natural language inference data}.
\newblock In \emph{Proceedings of the 2018 Conference of the North {A}merican
  Chapter of the Association for Computational Linguistics: Human Language
  Technologies, Volume 2 (Short Papers)}, pages 107--112, New Orleans,
  Louisiana. Association for Computational Linguistics.

\bibitem[{Hanselowski et~al.(2019)Hanselowski, Stab, Schulz, Li, and
  Gurevych}]{Hanselowski2019ARA}
Andreas Hanselowski, Christian Stab, Claudia Schulz, Zile Li, and Iryna
  Gurevych. 2019.
\newblock \href {https://doi.org/10.18653/v1/K19-1046} {A richly annotated
  corpus for different tasks in automated fact-checking}.
\newblock In \emph{Proceedings of the 23rd Conference on Computational Natural
  Language Learning (CoNLL)}, pages 493--503, Hong Kong, China. Association for
  Computational Linguistics.

\bibitem[{Hardalov et~al.(2021)Hardalov, Arora, Nakov, and
  Augenstein}]{Hardalov2021ASO}
Momchil Hardalov, Arnav Arora, Preslav Nakov, and Isabelle Augenstein. 2021.
\newblock A survey on stance detection for mis- and disinformation
  identification.
\newblock \emph{ArXiv}, abs/2103.00242.

\bibitem[{Hasanain et~al.(2019)Hasanain, Suwaileh, Elsayed,
  Barr{\'{o}}n{-}Cede{\~{n}}o, and Nakov}]{HasanainSEBN19}
Maram Hasanain, Reem Suwaileh, Tamer Elsayed, Alberto
  Barr{\'{o}}n{-}Cede{\~{n}}o, and Preslav Nakov. 2019.
\newblock \href {http://ceur-ws.org/Vol-2380/paper\_270.pdf} {Overview of the
  {CLEF-2019} {CheckThat!} lab: Automatic identification and verification of
  claims. task 2: Evidence and factuality}.
\newblock In \emph{Working Notes of {CLEF} 2019 - Conference and Labs of the
  Evaluation Forum, Lugano, Switzerland, September 9-12, 2019}, volume 2380 of
  \emph{{CEUR} Workshop Proceedings}. CEUR-WS.org.

\bibitem[{Hassan et~al.(2015)Hassan, Li, and Tremayne}]{hassan2015detecting}
Naeemul Hassan, Chengkai Li, and Mark Tremayne. 2015.
\newblock \href {https://doi.org/10.1145/2806416.2806652} {Detecting
  check-worthy factual claims in presidential debates}.
\newblock In \emph{Proceedings of the 24th {ACM} International Conference on
  Information and Knowledge Management, {CIKM} 2015, Melbourne, VIC, Australia,
  October 19 - 23, 2015}, pages 1835--1838. {ACM}.

\bibitem[{Hassan et~al.(2017)Hassan, Zhang, Arslan, Caraballo, Jimenez,
  Gawsane, Hasan, Joseph, Kulkarni, Nayak, Sable, Li, and
  Tremayne}]{HassanZACJGHJKN17}
Naeemul Hassan, Gensheng Zhang, Fatma Arslan, Josue Caraballo, Damian Jimenez,
  Siddhant Gawsane, Shohedul Hasan, Minumol Joseph, Aaditya Kulkarni,
  Anil~Kumar Nayak, Vikas Sable, Chengkai Li, and Mark Tremayne. 2017.
\newblock \href {https://doi.org/10.14778/3137765.3137815} {{ClaimBuster}: The
  first-ever end-to-end fact-checking system}.
\newblock \emph{Proceedings of the {VLDB} Endowment}, 10(12):1945--1948.

\bibitem[{Hermann et~al.(2015)Hermann, Kocisk{\'{y}}, Grefenstette, Espeholt,
  Kay, Suleyman, and Blunsom}]{HermannKGEKSB15}
Karl~Moritz Hermann, Tom{\'{a}}s Kocisk{\'{y}}, Edward Grefenstette, Lasse
  Espeholt, Will Kay, Mustafa Suleyman, and Phil Blunsom. 2015.
\newblock \href
  {https://proceedings.neurips.cc/paper/2015/hash/afdec7005cc9f14302cd0474fd0f3c96-Abstract.html}
  {Teaching machines to read and comprehend}.
\newblock In \emph{Advances in Neural Information Processing Systems 28: Annual
  Conference on Neural Information Processing Systems 2015, December 7-12,
  2015, Montreal, Quebec, Canada}, pages 1693--1701.

\bibitem[{Hochreiter and Schmidhuber(1997)}]{HochreiterS97}
Sepp Hochreiter and J{\"{u}}rgen Schmidhuber. 1997.
\newblock \href {https://doi.org/10.1162/neco.1997.9.8.1735} {Long short-term
  memory}.
\newblock \emph{Neural Computation}, 9(8):1735--1780.

\bibitem[{Horne et~al.(2018)Horne, Khedr, and Adali}]{HorneKA18}
Benjamin~D. Horne, Sara Khedr, and Sibel Adali. 2018.
\newblock \href {https://aaai.org/ocs/index.php/ICWSM/ICWSM18/paper/view/17796}
  {Sampling the news producers: {A} large news and feature data set for the
  study of the complex media landscape}.
\newblock In \emph{Proceedings of the Twelfth International Conference on Web
  and Social Media, {ICWSM} 2018, Stanford, California, USA, June 25-28, 2018},
  pages 518--527. {AAAI} Press.

\bibitem[{Hossain et~al.(2020)Hossain, Logan~IV, Ugarte, Matsubara, Young, and
  Singh}]{Hossain2020COVIDLiesDC}
Tamanna Hossain, Robert~L. Logan~IV, Arjuna Ugarte, Yoshitomo Matsubara, Sean
  Young, and Sameer Singh. 2020.
\newblock \href {https://doi.org/10.18653/v1/2020.nlpcovid19-2.11}
  {{COVIDL}ies: Detecting {COVID}-19 misinformation on social media}.
\newblock In \emph{Proceedings of the 1st Workshop on {NLP} for {COVID}-19
  (Part 2) at {EMNLP} 2020}, Online. Association for Computational Linguistics.

\bibitem[{Islam et~al.(2020)Islam, Liu, Wang, and Xu}]{IslamLWX20}
Md.~Rafiqul Islam, Shaowu Liu, Xianzhi Wang, and Guandong Xu. 2020.
\newblock \href {https://doi.org/10.1007/s13278-020-00696-x} {Deep learning for
  misinformation detection on online social networks: a survey and new
  perspectives}.
\newblock \emph{Soc. Netw. Anal. Min.}, 10(1):82.

\bibitem[{Jacovi and Goldberg(2020)}]{jacovi2020towards}
Alon Jacovi and Yoav Goldberg. 2020.
\newblock \href {https://doi.org/10.18653/v1/2020.acl-main.386} {Towards
  faithfully interpretable {NLP} systems: How should we define and evaluate
  faithfulness?}
\newblock In \emph{Proceedings of the 58th Annual Meeting of the Association
  for Computational Linguistics}, pages 4198--4205, Online. Association for
  Computational Linguistics.

\bibitem[{Jain and Wallace(2019)}]{Jain2019AttentionIN}
Sarthak Jain and Byron~C. Wallace. 2019.
\newblock \href {https://doi.org/10.18653/v1/N19-1357} {{A}ttention is not
  {E}xplanation}.
\newblock In \emph{Proceedings of the 2019 Conference of the North {A}merican
  Chapter of the Association for Computational Linguistics: Human Language
  Technologies, Volume 1 (Long and Short Papers)}, pages 3543--3556,
  Minneapolis, Minnesota. Association for Computational Linguistics.

\bibitem[{Jiang et~al.(2020)Jiang, Bordia, Zhong, Dognin, Singh, and
  Bansal}]{jiang2020hover}
Yichen Jiang, Shikha Bordia, Zheng Zhong, Charles Dognin, Maneesh Singh, and
  Mohit Bansal. 2020.
\newblock \href {https://doi.org/10.18653/v1/2020.findings-emnlp.309}
  {{H}o{V}er: A dataset for many-hop fact extraction and claim verification}.
\newblock In \emph{Findings of the Association for Computational Linguistics:
  EMNLP 2020}, pages 3441--3460, Online. Association for Computational
  Linguistics.

\bibitem[{Johnson et~al.(2017)Johnson, Douze, and
  J{\'{e}}gou}]{johnson2017faiss}
Jeff Johnson, Matthijs Douze, and Herv{\'{e}} J{\'{e}}gou. 2017.
\newblock \href {http://arxiv.org/abs/1702.08734} {Billion-scale similarity
  search with gpus}.
\newblock \emph{CoRR}, abs/1702.08734.

\bibitem[{Jowett and O'Donnell(2019)}]{jowett2019propaganda}
Garth~S. Jowett and Victoria O'Donnell. 2019.
\newblock \emph{Propaganda \& Persuasion}, 7th edition.
\newblock {SAGE} Publications.

\bibitem[{Karimi~Mahabadi et~al.(2020)Karimi~Mahabadi, Belinkov, and
  Henderson}]{karimi-mahabadi-etal-2020-end}
Rabeeh Karimi~Mahabadi, Yonatan Belinkov, and James Henderson. 2020.
\newblock \href {https://doi.org/10.18653/v1/2020.acl-main.769} {End-to-end
  bias mitigation by modelling biases in corpora}.
\newblock In \emph{Proceedings of the 58th Annual Meeting of the Association
  for Computational Linguistics}, pages 8706--8716, Online. Association for
  Computational Linguistics.

\bibitem[{Kazemi et~al.(2021)Kazemi, Garimella, Gaffney, and
  Hale}]{kazemi2021claim}
Ashkan Kazemi, Kiran Garimella, Devin Gaffney, and Scott Hale. 2021.
\newblock \href {https://doi.org/10.18653/v1/2021.acl-long.347} {Claim matching
  beyond {E}nglish to scale global fact-checking}.
\newblock In \emph{Proceedings of the 59th Annual Meeting of the Association
  for Computational Linguistics and the 11th International Joint Conference on
  Natural Language Processing (Volume 1: Long Papers)}, pages 4504--4517,
  Online. Association for Computational Linguistics.

\bibitem[{Khouja(2020)}]{Khouja2020StancePA}
Jude Khouja. 2020.
\newblock \href {https://doi.org/10.18653/v1/2020.fever-1.2} {Stance prediction
  and claim verification: An {A}rabic perspective}.
\newblock In \emph{Proceedings of the Third Workshop on Fact Extraction and
  VERification (FEVER)}, pages 8--17, Online. Association for Computational
  Linguistics.

\bibitem[{Kilicoglu et~al.(2012)Kilicoglu, Shin, Fiszman, Rosemblat, and
  Rindflesch}]{KilicogluSFRR12}
Halil Kilicoglu, Dongwook Shin, Marcelo Fiszman, Graciela Rosemblat, and
  Thomas~C. Rindflesch. 2012.
\newblock \href {https://doi.org/10.1093/bioinformatics/bts591} {{SemMedDB}: a
  pubmed-scale repository of biomedical semantic predications}.
\newblock \emph{Bioinform.}, 28(23):3158--3160.

\bibitem[{Kim and Choi(2020)}]{kim-choi-2020-unsupervised}
Jiseong Kim and Key-sun Choi. 2020.
\newblock \href {https://doi.org/10.18653/v1/2020.coling-main.147}
  {Unsupervised fact checking by counter-weighted positive and negative
  evidential paths in a knowledge graph}.
\newblock In \emph{Proceedings of the 28th International Conference on
  Computational Linguistics}, pages 1677--1686, Barcelona, Spain (Online).
  International Committee on Computational Linguistics.

\bibitem[{Kipf and Welling(2017)}]{Kipf2017SemiSupervisedCW}
Thomas~N. Kipf and Max Welling. 2017.
\newblock \href {https://openreview.net/forum?id=SJU4ayYgl} {Semi-supervised
  classification with graph convolutional networks}.
\newblock In \emph{5th International Conference on Learning Representations,
  {ICLR} 2017, Toulon, France, April 24-26, 2017, Conference Track
  Proceedings}. OpenReview.net.

\bibitem[{Kochkina et~al.(2017)Kochkina, Liakata, and
  Augenstein}]{Kochkina2017TuringAS}
Elena Kochkina, Maria Liakata, and Isabelle Augenstein. 2017.
\newblock \href {https://doi.org/10.18653/v1/S17-2083} {{T}uring at
  {S}em{E}val-2017 task 8: Sequential approach to rumour stance classification
  with branch-{LSTM}}.
\newblock In \emph{Proceedings of the 11th International Workshop on Semantic
  Evaluation ({S}em{E}val-2017)}, pages 475--480, Vancouver, Canada.
  Association for Computational Linguistics.

\bibitem[{Konstantinovskiy et~al.(2021)Konstantinovskiy, Price, Babakar, and
  Zubiaga}]{konstantinovskiy2018towards}
Lev Konstantinovskiy, Oliver Price, Mevan Babakar, and Arkaitz Zubiaga. 2021.
\newblock Toward automated factchecking: Developing an annotation schema and
  benchmark for consistent automated claim detection.
\newblock \emph{Digital Threats: Research and Practice}, 2(2):1--16.

\bibitem[{Kotonya and Toni(2020{\natexlab{a}})}]{Kotonya2020ExplainableAF}
Neema Kotonya and Francesca Toni. 2020{\natexlab{a}}.
\newblock \href {https://doi.org/10.18653/v1/2020.coling-main.474} {Explainable
  automated fact-checking: A survey}.
\newblock In \emph{Proceedings of the 28th International Conference on
  Computational Linguistics}, pages 5430--5443, Barcelona, Spain (Online).
  International Committee on Computational Linguistics.

\bibitem[{Kotonya and Toni(2020{\natexlab{b}})}]{kotonya2020explainable}
Neema Kotonya and Francesca Toni. 2020{\natexlab{b}}.
\newblock \href {https://doi.org/10.18653/v1/2020.emnlp-main.623} {Explainable
  automated fact-checking for public health claims}.
\newblock In \emph{Proceedings of the 2020 Conference on Empirical Methods in
  Natural Language Processing (EMNLP)}, pages 7740--7754, Online. Association
  for Computational Linguistics.

\bibitem[{K{\"{u}}{\c{c}}{\"{u}}k and Can(2020)}]{Kk2020StanceD}
Dilek K{\"{u}}{\c{c}}{\"{u}}k and Fazli Can. 2020.
\newblock \href {https://doi.org/10.1145/3369026} {Stance detection: {A}
  survey}.
\newblock \emph{{ACM} Computing Surveys}, 53(1):12:1--12:37.

\bibitem[{Lazer et~al.(2018)Lazer, Baum, Benkler, Berinsky, Greenhill, Menczer,
  Metzger, Nyhan, Pennycook, Rothschild et~al.}]{lazer2018science}
David~MJ Lazer, Matthew~A Baum, Yochai Benkler, Adam~J Berinsky, Kelly~M
  Greenhill, Filippo Menczer, Miriam~J Metzger, Brendan Nyhan, Gordon
  Pennycook, David Rothschild, et~al. 2018.
\newblock The science of fake news.
\newblock \emph{Science}, 359(6380):1094--1096.

\bibitem[{Lee et~al.(2021)Lee, Bang, Madotto, and Fung}]{lee-etal-2021-towards}
Nayeon Lee, Yejin Bang, Andrea Madotto, and Pascale Fung. 2021.
\newblock \href {https://doi.org/10.18653/v1/2021.naacl-main.158} {Towards
  few-shot fact-checking via perplexity}.
\newblock In \emph{Proceedings of the 2021 Conference of the North American
  Chapter of the Association for Computational Linguistics: Human Language
  Technologies}, pages 1971--1981, Online. Association for Computational
  Linguistics.

\bibitem[{Lee et~al.(2020)Lee, Li, Wang, Yih, Ma, and
  Khabsa}]{lee-etal-2020-language}
Nayeon Lee, Belinda~Z. Li, Sinong Wang, Wen-tau Yih, Hao Ma, and Madian Khabsa.
  2020.
\newblock \href {https://doi.org/10.18653/v1/2020.fever-1.5} {Language models
  as fact checkers?}
\newblock In \emph{Proceedings of the Third Workshop on Fact Extraction and
  VERification (FEVER)}, pages 36--41, Online. Association for Computational
  Linguistics.

\bibitem[{Lertvittayakumjorn and
  Toni(2019)}]{lertvittayakumjorn-toni-2019-human}
Piyawat Lertvittayakumjorn and Francesca Toni. 2019.
\newblock \href {https://doi.org/10.18653/v1/D19-1523} {Human-grounded
  evaluations of explanation methods for text classification}.
\newblock In \emph{Proceedings of the 2019 Conference on Empirical Methods in
  Natural Language Processing and the 9th International Joint Conference on
  Natural Language Processing (EMNLP-IJCNLP)}, pages 5195--5205, Hong Kong,
  China. Association for Computational Linguistics.

\bibitem[{Lewandowsky et~al.(2012)Lewandowsky, Ecker, Seifert, Schwarz, and
  Cook}]{Lewandowsky2012}
Stephan Lewandowsky, Ullrich~K.H. Ecker, Colleen~M. Seifert, Norbert Schwarz,
  and John Cook. 2012.
\newblock \href {https://doi.org/10.1177/1529100612451018} {{Misinformation and
  Its Correction: Continued Influence and Successful Debiasing}}.
\newblock \emph{Psychological Science in the Public Interest, Supplement},
  13(3):106--131.

\bibitem[{Lewandowsky and van~der Linden(2021)}]{lewandosky2021prebunking}
Stephan Lewandowsky and Sander van~der Linden. 2021.
\newblock \href {https://doi.org/10.1080/10463283.2021.1876983} {Countering
  misinformation and fake news through inoculation and prebunking}.
\newblock \emph{European Review of Social Psychology}, 0(0):1--38.

\bibitem[{Lewis et~al.(2008)Lewis, Williams, Franklin, Thomas, and
  Mosdell}]{lewis2008quality}
Justin Matthew~Wren Lewis, Andy Williams, Robert~Arthur Franklin, James Thomas,
  and Nicholas~Alexander Mosdell. 2008.
\newblock The quality and independence of british journalism.
\newblock \emph{Mediawise}.

\bibitem[{Lewis et~al.(2020)Lewis, Perez, Piktus, Petroni, Karpukhin, Goyal,
  K{\"{u}}ttler, Lewis, Yih, Rockt{\"{a}}schel, Riedel, and
  Kiela}]{lewis2020retrieval}
Patrick S.~H. Lewis, Ethan Perez, Aleksandra Piktus, Fabio Petroni, Vladimir
  Karpukhin, Naman Goyal, Heinrich K{\"{u}}ttler, Mike Lewis, Wen{-}tau Yih,
  Tim Rockt{\"{a}}schel, Sebastian Riedel, and Douwe Kiela. 2020.
\newblock \href
  {https://proceedings.neurips.cc/paper/2020/hash/6b493230205f780e1bc26945df7481e5-Abstract.html}
  {Retrieval-augmented generation for knowledge-intensive {NLP} tasks}.
\newblock In \emph{Advances in Neural Information Processing Systems 33: Annual
  Conference on Neural Information Processing Systems 2020, NeurIPS 2020,
  December 6-12, 2020, virtual}.

\bibitem[{Li et~al.(2020)Li, Sujana, and Kao}]{li-etal-2020-exploiting}
Jiawen Li, Yudianto Sujana, and Hung-Yu Kao. 2020.
\newblock \href {https://doi.org/10.18653/v1/2020.coling-main.473} {Exploiting
  microblog conversation structures to detect rumors}.
\newblock In \emph{Proceedings of the 28th International Conference on
  Computational Linguistics}, pages 5420--5429, Barcelona, Spain (Online).
  International Committee on Computational Linguistics.

\bibitem[{Li et~al.(2015)Li, Gao, Meng, Li, Su, Zhao, Fan, and
  Han}]{li2016survey}
Yaliang Li, Jing Gao, Chuishi Meng, Qi~Li, Lu~Su, Bo~Zhao, Wei Fan, and Jiawei
  Han. 2015.
\newblock \href {https://doi.org/10.1145/2897350.2897352} {A survey on truth
  discovery}.
\newblock \emph{{SIGKDD} Explor.}, 17(2):1--16.

\bibitem[{Lillie et~al.(2019)Lillie, Middelboe, and
  Derczynski}]{Lillie2019JointRS}
Anders~Edelbo Lillie, Emil~Refsgaard Middelboe, and Leon Derczynski. 2019.
\newblock \href {https://www.aclweb.org/anthology/W19-6122} {Joint rumour
  stance and veracity prediction}.
\newblock In \emph{Proceedings of the 22nd Nordic Conference on Computational
  Linguistics}, pages 208--221, Turku, Finland. Link{\"o}ping University
  Electronic Press.

\bibitem[{van~der Linden et~al.(2017)van~der Linden, Leiserowitz, Rosenthal,
  and Maibach}]{vanderlinden2017inoculating}
Sander van~der Linden, Anthony Leiserowitz, Seth Rosenthal, and Edward Maibach.
  2017.
\newblock \href {https://doi.org/https://doi.org/10.1002/gch2.201600008}
  {Inoculating the public against misinformation about climate change}.
\newblock \emph{Global Challenges}, 1(2):1600008.

\bibitem[{Lipton(2018)}]{Lipton18}
Zachary~C. Lipton. 2018.
\newblock \href {https://doi.org/10.1145/3233231} {The mythos of model
  interpretability}.
\newblock \emph{Commun. {ACM}}, 61(10):36--43.

\bibitem[{Liu et~al.(2020)Liu, Xiong, Sun, and Liu}]{Liu2020KernelGA}
Zhenghao Liu, Chenyan Xiong, Maosong Sun, and Zhiyuan Liu. 2020.
\newblock \href {https://doi.org/10.18653/v1/2020.acl-main.655} {Fine-grained
  fact verification with kernel graph attention network}.
\newblock In \emph{Proceedings of the 58th Annual Meeting of the Association
  for Computational Linguistics}, pages 7342--7351, Online. Association for
  Computational Linguistics.

\bibitem[{Lu and Li(2020)}]{Lu2020GCANGC}
Yi-Ju Lu and Cheng-Te Li. 2020.
\newblock \href {https://doi.org/10.18653/v1/2020.acl-main.48} {{GCAN}:
  Graph-aware co-attention networks for explainable fake news detection on
  social media}.
\newblock In \emph{Proceedings of the 58th Annual Meeting of the Association
  for Computational Linguistics}, pages 505--514, Online. Association for
  Computational Linguistics.

\bibitem[{Luken et~al.(2018)Luken, Jiang, and
  de~Marneffe}]{luken-etal-2018-qed}
Jackson Luken, Nanjiang Jiang, and Marie-Catherine de~Marneffe. 2018.
\newblock \href {https://doi.org/10.18653/v1/W18-5526} {{QED}: A fact
  verification system for the {FEVER} shared task}.
\newblock In \emph{Proceedings of the First Workshop on Fact Extraction and
  {VER}ification ({FEVER})}, pages 156--160, Brussels, Belgium. Association for
  Computational Linguistics.

\bibitem[{Ma et~al.(2019)Ma, Gao, Joty, and Wong}]{ma-etal-2019-sentence}
Jing Ma, Wei Gao, Shafiq Joty, and Kam-Fai Wong. 2019.
\newblock \href {https://doi.org/10.18653/v1/P19-1244} {Sentence-level evidence
  embedding for claim verification with hierarchical attention networks}.
\newblock In \emph{Proceedings of the 57th Annual Meeting of the Association
  for Computational Linguistics}, pages 2561--2571, Florence, Italy.
  Association for Computational Linguistics.

\bibitem[{Ma et~al.(2016)Ma, Gao, Mitra, Kwon, Jansen, Wong, and
  Cha}]{Ma2016DetectingRF}
Jing Ma, Wei Gao, Prasenjit Mitra, Sejeong Kwon, Bernard~J. Jansen, Kam{-}Fai
  Wong, and Meeyoung Cha. 2016.
\newblock \href {http://www.ijcai.org/Abstract/16/537} {Detecting rumors from
  microblogs with recurrent neural networks}.
\newblock In \emph{Proceedings of the Twenty-Fifth International Joint
  Conference on Artificial Intelligence, {IJCAI} 2016, New York, NY, USA, 9-15
  July 2016}, pages 3818--3824. {IJCAI/AAAI} Press.

\bibitem[{Ma et~al.(2018)Ma, Gao, and Wong}]{Ma2018RumorDO}
Jing Ma, Wei Gao, and Kam-Fai Wong. 2018.
\newblock \href {https://doi.org/10.18653/v1/P18-1184} {Rumor detection on
  {T}witter with tree-structured recursive neural networks}.
\newblock In \emph{Proceedings of the 56th Annual Meeting of the Association
  for Computational Linguistics (Volume 1: Long Papers)}, pages 1980--1989,
  Melbourne, Australia. Association for Computational Linguistics.

\bibitem[{Maillard et~al.(2021)Maillard, Karpukhin, Petroni, Yih, Oguz,
  Stoyanov, and Ghosh}]{maillard-etal-2021-multi}
Jean Maillard, Vladimir Karpukhin, Fabio Petroni, Wen-tau Yih, Barlas Oguz,
  Veselin Stoyanov, and Gargi Ghosh. 2021.
\newblock \href {https://doi.org/10.18653/v1/2021.acl-long.89} {Multi-task
  retrieval for knowledge-intensive tasks}.
\newblock In \emph{Proceedings of the 59th Annual Meeting of the Association
  for Computational Linguistics and the 11th International Joint Conference on
  Natural Language Processing (Volume 1: Long Papers)}, pages 1098--1111,
  Online. Association for Computational Linguistics.

\bibitem[{Maynez et~al.(2020)Maynez, Narayan, Bohnet, and
  McDonald}]{maynez-etal-2020-faithfulness}
Joshua Maynez, Shashi Narayan, Bernd Bohnet, and Ryan McDonald. 2020.
\newblock \href {https://doi.org/10.18653/v1/2020.acl-main.173} {On
  faithfulness and factuality in abstractive summarization}.
\newblock In \emph{Proceedings of the 58th Annual Meeting of the Association
  for Computational Linguistics}, pages 1906--1919, Online. Association for
  Computational Linguistics.

\bibitem[{McCoy et~al.(2019)McCoy, Pavlick, and Linzen}]{McCoy2019RightFT}
Tom McCoy, Ellie Pavlick, and Tal Linzen. 2019.
\newblock \href {https://doi.org/10.18653/v1/P19-1334} {Right for the wrong
  reasons: Diagnosing syntactic heuristics in natural language inference}.
\newblock In \emph{Proceedings of the 57th Annual Meeting of the Association
  for Computational Linguistics}, pages 3428--3448, Florence, Italy.
  Association for Computational Linguistics.

\bibitem[{Mena(2019)}]{mena2019principles}
Paul Mena. 2019.
\newblock Principles and boundaries of fact-checking: Journalists’
  perceptions.
\newblock \emph{Journalism Practice}, 13(6):657--672.

\bibitem[{Mihalcea and Strapparava(2009)}]{mihalcea2009lie}
Rada Mihalcea and Carlo Strapparava. 2009.
\newblock \href {https://www.aclweb.org/anthology/P09-2078} {The lie detector:
  Explorations in the automatic recognition of deceptive language}.
\newblock In \emph{Proceedings of the {ACL}-{IJCNLP} 2009 Conference Short
  Papers}, pages 309--312, Suntec, Singapore. Association for Computational
  Linguistics.

\bibitem[{Mihaylova et~al.(2018)Mihaylova, Nakov, M{\`{a}}rquez,
  Barr{\'{o}}n{-}Cede{\~{n}}o, Mohtarami, Karadzhov, and
  Glass}]{Mihaylova2018FactCI}
Tsvetomila Mihaylova, Preslav Nakov, Llu{\'{\i}}s M{\`{a}}rquez, Alberto
  Barr{\'{o}}n{-}Cede{\~{n}}o, Mitra Mohtarami, Georgi Karadzhov, and James~R.
  Glass. 2018.
\newblock \href
  {https://www.aaai.org/ocs/index.php/AAAI/AAAI18/paper/view/16780} {Fact
  checking in community forums}.
\newblock In \emph{Proceedings of the Thirty-Second {AAAI} Conference on
  Artificial Intelligence, (AAAI-18), the 30th innovative Applications of
  Artificial Intelligence (IAAI-18), and the 8th {AAAI} Symposium on
  Educational Advances in Artificial Intelligence (EAAI-18), New Orleans,
  Louisiana, USA, February 2-7, 2018}, pages 5309--5316. {AAAI} Press.

\bibitem[{Mitra and Gilbert(2015)}]{MitraG15}
Tanushree Mitra and Eric Gilbert. 2015.
\newblock \href
  {http://www.aaai.org/ocs/index.php/ICWSM/ICWSM15/paper/view/10582}
  {{CREDBANK:} {A} large-scale social media corpus with associated credibility
  annotations}.
\newblock In \emph{Proceedings of the Ninth International Conference on Web and
  Social Media, {ICWSM} 2015, University of Oxford, Oxford, UK, May 26-29,
  2015}, pages 258--267. {AAAI} Press.

\bibitem[{Monti et~al.(2019)Monti, Frasca, Eynard, Mannion, and
  Bronstein}]{monti2019fake}
Federico Monti, Fabrizio Frasca, Davide Eynard, Damon Mannion, and Michael~M.
  Bronstein. 2019.
\newblock \href {http://arxiv.org/abs/1902.06673} {Fake news detection on
  social media using geometric deep learning}.
\newblock \emph{CoRR}, abs/1902.06673.

\bibitem[{Mu and Aletras(2020)}]{Mu2020IdentifyingTU}
Yida Mu and Nikolaos Aletras. 2020.
\newblock Identifying twitter users who repost unreliable news sources with
  linguistic information.
\newblock \emph{PeerJ Computer Science}, 6.

\bibitem[{Nakamura et~al.(2020)Nakamura, Levy, and Wang}]{NakamuraLW20}
Kai Nakamura, Sharon Levy, and William~Yang Wang. 2020.
\newblock \href {https://www.aclweb.org/anthology/2020.lrec-1.755/} {Fakeddit:
  {A} new multimodal benchmark dataset for fine-grained fake news detection}.
\newblock In \emph{Proceedings of The 12th Language Resources and Evaluation
  Conference, {LREC} 2020, Marseille, France, May 11-16, 2020}, pages
  6149--6157. European Language Resources Association.

\bibitem[{Nakashole and Mitchell(2014)}]{Nakashole2014LanguageAwareTA}
Ndapandula Nakashole and Tom~M. Mitchell. 2014.
\newblock \href {https://doi.org/10.3115/v1/P14-1095} {Language-aware truth
  assessment of fact candidates}.
\newblock In \emph{Proceedings of the 52nd Annual Meeting of the Association
  for Computational Linguistics (Volume 1: Long Papers)}, pages 1009--1019,
  Baltimore, Maryland. Association for Computational Linguistics.

\bibitem[{Nakov(2020)}]{nakov2020can}
Preslav Nakov. 2020.
\newblock \href {http://arxiv.org/abs/2008.04374} {Can we spot the "fake news"
  before it was even written?}
\newblock \emph{CoRR}, abs/2008.04374.

\bibitem[{Nakov et~al.(2021{\natexlab{a}})Nakov, Corney, Hasanain, Alam,
  Elsayed, Barr{\'{o}}n{-}Cede{\~{n}}o, Papotti, Shaar, and
  Martino}]{nakov2021}
Preslav Nakov, David P.~A. Corney, Maram Hasanain, Firoj Alam, Tamer Elsayed,
  Alberto Barr{\'{o}}n{-}Cede{\~{n}}o, Paolo Papotti, Shaden Shaar, and
  Giovanni Da~San Martino. 2021{\natexlab{a}}.
\newblock \href {http://arxiv.org/abs/2103.07769} {Automated fact-checking for
  assisting human fact-checkers}.
\newblock \emph{CoRR}, abs/2103.07769.

\bibitem[{Nakov et~al.(2021{\natexlab{b}})Nakov, Martino, Elsayed,
  Barr{\'{o}}n{-}Cede{\~{n}}o, M{\'{\i}}guez, Shaar, Alam, Haouari, Hasanain,
  Babulkov, Nikolov, Shahi, Stru{\ss}, and Mandl}]{NakovMEBMSAHHBN21}
Preslav Nakov, Giovanni Da~San Martino, Tamer Elsayed, Alberto
  Barr{\'{o}}n{-}Cede{\~{n}}o, Rub{\'{e}}n M{\'{\i}}guez, Shaden Shaar, Firoj
  Alam, Fatima Haouari, Maram Hasanain, Nikolay Babulkov, Alex Nikolov,
  Gautam~Kishore Shahi, Julia~Maria Stru{\ss}, and Thomas Mandl.
  2021{\natexlab{b}}.
\newblock \href {https://doi.org/10.1007/978-3-030-72240-1\_75} {The
  {CLEF-2021} {CheckThat!} lab on detecting check-worthy claims, previously
  fact-checked claims, and fake news}.
\newblock In \emph{Advances in Information Retrieval - 43rd European Conference
  on {IR} Research, {ECIR} 2021, Virtual Event, March 28 - April 1, 2021,
  Proceedings, Part {II}}, volume 12657 of \emph{Lecture Notes in Computer
  Science}, pages 639--649. Springer.

\bibitem[{Nam et~al.(2018)Nam, Kim, Kim, Jung, Han, and Choi}]{NamKKJHC18}
Sangha Nam, Eun{-}Kyung Kim, Jiho Kim, Yoosung Jung, Kijong Han, and Key{-}Sun
  Choi. 2018.
\newblock \href {https://www.aclweb.org/anthology/C18-2005/} {A korean
  knowledge extraction system for enriching a kbox}.
\newblock In \emph{{COLING} 2018, The 27th International Conference on
  Computational Linguistics: System Demonstrations, Santa Fe, New Mexico,
  August 20-26, 2018}, pages 20--24. Association for Computational Linguistics.

\bibitem[{Nie et~al.(2019{\natexlab{a}})Nie, Chen, and
  Bansal}]{Nie2019CombiningFE}
Yixin Nie, Haonan Chen, and Mohit Bansal. 2019{\natexlab{a}}.
\newblock \href {https://doi.org/10.1609/aaai.v33i01.33016859} {Combining fact
  extraction and verification with neural semantic matching networks}.
\newblock In \emph{The Thirty-Third {AAAI} Conference on Artificial
  Intelligence, {AAAI} 2019, The Thirty-First Innovative Applications of
  Artificial Intelligence Conference, {IAAI} 2019, The Ninth {AAAI} Symposium
  on Educational Advances in Artificial Intelligence, {EAAI} 2019, Honolulu,
  Hawaii, USA, January 27 - February 1, 2019}, pages 6859--6866. {AAAI} Press.

\bibitem[{Nie et~al.(2019{\natexlab{b}})Nie, Wang, and
  Bansal}]{nie-etal-2019-revealing}
Yixin Nie, Songhe Wang, and Mohit Bansal. 2019{\natexlab{b}}.
\newblock \href {https://doi.org/10.18653/v1/D19-1258} {Revealing the
  importance of semantic retrieval for machine reading at scale}.
\newblock In \emph{Proceedings of the 2019 Conference on Empirical Methods in
  Natural Language Processing and the 9th International Joint Conference on
  Natural Language Processing (EMNLP-IJCNLP)}, pages 2553--2566, Hong Kong,
  China. Association for Computational Linguistics.

\bibitem[{N{\o}rregaard and Derczynski(2021)}]{NorregaardD21}
Jeppe N{\o}rregaard and Leon Derczynski. 2021.
\newblock \href {https://www.aclweb.org/anthology/2021.nodalida-main.47/}
  {{DanFEVER}: claim verification dataset for danish}.
\newblock In \emph{Proceedings of the 23rd Nordic Conference on Computational
  Linguistics, NoDaLiDa 2021, Reykjavik, Iceland (Online), May 31 - June 2,
  2021}, pages 422--428. Link{\"{o}}ping University Electronic Press, Sweden.

\bibitem[{N{\o}rregaard et~al.(2019)N{\o}rregaard, Horne, and
  Adali}]{NorregaardHA19}
Jeppe N{\o}rregaard, Benjamin~D. Horne, and Sibel Adali. 2019.
\newblock \href {https://aaai.org/ojs/index.php/ICWSM/article/view/3261}
  {{NELA-GT-2018:} {A} large multi-labelled news dataset for the study of
  misinformation in news articles}.
\newblock In \emph{Proceedings of the Thirteenth International Conference on
  Web and Social Media, {ICWSM} 2019, Munich, Germany, June 11-14, 2019}, pages
  630--638. {AAAI} Press.

\bibitem[{O'Neil(2016)}]{oneil2016weapons}
Cathy O'Neil. 2016.
\newblock \emph{Weapons of Math Destruction: How Big Data Increases Inequality
  and Threatens Democracy}.
\newblock Crown Publishing Group, USA.

\bibitem[{Oshikawa et~al.(2020)Oshikawa, Qian, and Wang}]{Oshikawa2020ASO}
Ray Oshikawa, Jing Qian, and William~Yang Wang. 2020.
\newblock \href {https://www.aclweb.org/anthology/2020.lrec-1.747} {A survey on
  natural language processing for fake news detection}.
\newblock In \emph{Proceedings of the 12th Language Resources and Evaluation
  Conference}, pages 6086--6093, Marseille, France. European Language Resources
  Association.

\bibitem[{Pan et~al.(2021)Pan, Chen, Xiong, Kan, and Wang}]{PanCXKW20}
Liangming Pan, Wenhu Chen, Wenhan Xiong, Min{-}Yen Kan, and William~Yang Wang.
  2021.
\newblock \href {https://doi.org/10.18653/v1/2021.acl-short.61} {Zero-shot fact
  verification by claim generation}.
\newblock In \emph{Proceedings of the 59th Annual Meeting of the Association
  for Computational Linguistics and the 11th International Joint Conference on
  Natural Language Processing, {ACL/IJCNLP} 2021, (Volume 2: Short Papers),
  Virtual Event, August 1-6, 2021}, pages 476--483. Association for
  Computational Linguistics.

\bibitem[{P{\'e}rez-Rosas et~al.(2018)P{\'e}rez-Rosas, Kleinberg, Lefevre, and
  Mihalcea}]{PrezRosas2018AutomaticDO}
Ver{\'o}nica P{\'e}rez-Rosas, Bennett Kleinberg, Alexandra Lefevre, and Rada
  Mihalcea. 2018.
\newblock \href {https://www.aclweb.org/anthology/C18-1287} {Automatic
  detection of fake news}.
\newblock In \emph{Proceedings of the 27th International Conference on
  Computational Linguistics}, pages 3391--3401, Santa Fe, New Mexico, USA.
  Association for Computational Linguistics.

\bibitem[{Petroni et~al.(2019)Petroni, Rockt{\"a}schel, Riedel, Lewis, Bakhtin,
  Wu, and Miller}]{petroni-etal-2019-language}
Fabio Petroni, Tim Rockt{\"a}schel, Sebastian Riedel, Patrick Lewis, Anton
  Bakhtin, Yuxiang Wu, and Alexander Miller. 2019.
\newblock \href {https://doi.org/10.18653/v1/D19-1250} {Language models as
  knowledge bases?}
\newblock In \emph{Proceedings of the 2019 Conference on Empirical Methods in
  Natural Language Processing and the 9th International Joint Conference on
  Natural Language Processing (EMNLP-IJCNLP)}, pages 2463--2473, Hong Kong,
  China. Association for Computational Linguistics.

\bibitem[{Poliak et~al.(2018)Poliak, Naradowsky, Haldar, Rudinger, and
  Van~Durme}]{poliak-etal-2018-hypothesis}
Adam Poliak, Jason Naradowsky, Aparajita Haldar, Rachel Rudinger, and Benjamin
  Van~Durme. 2018.
\newblock \href {https://doi.org/10.18653/v1/S18-2023} {Hypothesis only
  baselines in natural language inference}.
\newblock In \emph{Proceedings of the Seventh Joint Conference on Lexical and
  Computational Semantics}, pages 180--191, New Orleans, Louisiana. Association
  for Computational Linguistics.

\bibitem[{Pomerleau and Rao(2017)}]{pomerleau2017fake}
Dean Pomerleau and Delip Rao. 2017.
\newblock The fake news challenge: Exploring how artificial intelligence
  technologies could be leveraged to combat fake news.
\newblock \emph{Fake News Challenge}.

\bibitem[{Popat et~al.(2016)Popat, Mukherjee, Str{\"{o}}tgen, and
  Weikum}]{Popat2016CredibilityAO}
Kashyap Popat, Subhabrata Mukherjee, Jannik Str{\"{o}}tgen, and Gerhard Weikum.
  2016.
\newblock \href {https://doi.org/10.1145/2983323.2983661} {Credibility
  assessment of textual claims on the web}.
\newblock In \emph{Proceedings of the 25th {ACM} International Conference on
  Information and Knowledge Management, {CIKM} 2016, Indianapolis, IN, USA,
  October 24-28, 2016}, pages 2173--2178. {ACM}.

\bibitem[{Popat et~al.(2018)Popat, Mukherjee, Yates, and
  Weikum}]{Popat2018DeClarEDF}
Kashyap Popat, Subhabrata Mukherjee, Andrew Yates, and Gerhard Weikum. 2018.
\newblock \href {https://doi.org/10.18653/v1/D18-1003} {{D}e{C}lar{E}:
  Debunking fake news and false claims using evidence-aware deep learning}.
\newblock In \emph{Proceedings of the 2018 Conference on Empirical Methods in
  Natural Language Processing}, pages 22--32, Brussels, Belgium. Association
  for Computational Linguistics.

\bibitem[{Potthast et~al.(2018)Potthast, Kiesel, Reinartz, Bevendorff, and
  Stein}]{Potthast2018ASI}
Martin Potthast, Johannes Kiesel, Kevin Reinartz, Janek Bevendorff, and Benno
  Stein. 2018.
\newblock \href {https://doi.org/10.18653/v1/P18-1022} {A stylometric inquiry
  into hyperpartisan and fake news}.
\newblock In \emph{Proceedings of the 56th Annual Meeting of the Association
  for Computational Linguistics (Volume 1: Long Papers)}, pages 231--240,
  Melbourne, Australia. Association for Computational Linguistics.

\bibitem[{Pruthi et~al.(2020)Pruthi, Gupta, Dhingra, Neubig, and
  Lipton}]{Pruthi2020LearningTD}
Danish Pruthi, Mansi Gupta, Bhuwan Dhingra, Graham Neubig, and Zachary~C.
  Lipton. 2020.
\newblock \href {https://doi.org/10.18653/v1/2020.acl-main.432} {Learning to
  deceive with attention-based explanations}.
\newblock In \emph{Proceedings of the 58th Annual Meeting of the Association
  for Computational Linguistics}, pages 4782--4793, Online. Association for
  Computational Linguistics.

\bibitem[{Qazvinian et~al.(2011)Qazvinian, Rosengren, Radev, and
  Mei}]{Qazvinian2011RumorHI}
Vahed Qazvinian, Emily Rosengren, Dragomir~R. Radev, and Qiaozhu Mei. 2011.
\newblock \href {https://www.aclweb.org/anthology/D11-1147} {Rumor has it:
  Identifying misinformation in microblogs}.
\newblock In \emph{Proceedings of the 2011 Conference on Empirical Methods in
  Natural Language Processing}, pages 1589--1599, Edinburgh, Scotland, UK.
  Association for Computational Linguistics.

\bibitem[{Radford et~al.(2019)Radford, Wu, Child, Luan, Amodei, and
  Sutskever}]{radford2019language}
Alec Radford, Jeffrey Wu, Rewon Child, David Luan, Dario Amodei, and Ilya
  Sutskever. 2019.
\newblock Language models are unsupervised multitask learners.
\newblock \emph{OpenAI blog}, 1(8):9.

\bibitem[{Rashkin et~al.(2017)Rashkin, Choi, Jang, Volkova, and
  Choi}]{Rashkin2017TruthOV}
Hannah Rashkin, Eunsol Choi, Jin~Yea Jang, Svitlana Volkova, and Yejin Choi.
  2017.
\newblock \href {https://doi.org/10.18653/v1/D17-1317} {Truth of varying
  shades: Analyzing language in fake news and political fact-checking}.
\newblock In \emph{Proceedings of the 2017 Conference on Empirical Methods in
  Natural Language Processing}, pages 2931--2937, Copenhagen, Denmark.
  Association for Computational Linguistics.

\bibitem[{Redi et~al.(2019)Redi, Fetahu, Morgan, and Taraborelli}]{redi2019}
Miriam Redi, Besnik Fetahu, Jonathan~T. Morgan, and Dario Taraborelli. 2019.
\newblock \href {https://doi.org/10.1145/3308558.3313618} {{Citation Needed}:
  {A} taxonomy and algorithmic assessment of wikipedia's verifiability}.
\newblock In \emph{The World Wide Web Conference, {WWW} 2019, San Francisco,
  CA, USA, May 13-17, 2019}, pages 1567--1578. {ACM}.

\bibitem[{Roozenbeek and van~der Linden(2019)}]{roozenbeek2019game}
Jon Roozenbeek and Sander van~der Linden. 2019.
\newblock \href {https://doi.org/10.1080/13669877.2018.1443491} {The fake news
  game: actively inoculating against the risk of misinformation}.
\newblock \emph{Journal of Risk Research}, 22(5):570--580.

\bibitem[{Roozenbeek et~al.(2020)Roozenbeek, van~der Linden, and
  Nygren}]{roozenbeek2020pre}
Jon Roozenbeek, Sander van~der Linden, and Thomas Nygren. 2020.
\newblock \href {http://arxiv.org/abs/https://doi.org/10.37016/mr-2020-008}
  {Prebunking interventions based on the psychological theory of "inoculation"
  can reduce susceptibility to misinformation across cultures}.
\newblock \emph{The Harvard Kennedy School Misinformation Review}, 1(2).

\bibitem[{Saakyan et~al.(2021)Saakyan, Chakrabarty, and Muresan}]{SaakyanCM20}
Arkadiy Saakyan, Tuhin Chakrabarty, and Smaranda Muresan. 2021.
\newblock \href {https://doi.org/10.18653/v1/2021.acl-long.165} {{COVID-Fact}:
  Fact extraction and verification of real-world claims on {COVID-19}
  pandemic}.
\newblock In \emph{Proceedings of the 59th Annual Meeting of the Association
  for Computational Linguistics and the 11th International Joint Conference on
  Natural Language Processing, {ACL/IJCNLP} 2021, (Volume 1: Long Papers),
  Virtual Event, August 1-6, 2021}, pages 2116--2129. Association for
  Computational Linguistics.

\bibitem[{Salem et~al.(2019)Salem, Feel, Elbassuoni, Jaber, and
  Farah}]{SalemFEJF19}
Fatima K.~Abu Salem, Roaa~Al Feel, Shady Elbassuoni, Mohamad Jaber, and May
  Farah. 2019.
\newblock \href {https://aaai.org/ojs/index.php/ICWSM/article/view/3254}
  {{FA-KES:} {A} fake news dataset around the syrian war}.
\newblock In \emph{Proceedings of the Thirteenth International Conference on
  Web and Social Media, {ICWSM} 2019, Munich, Germany, June 11-14, 2019}, pages
  573--582. {AAAI} Press.

\bibitem[{Santia and Williams(2018)}]{SantiaW18}
Giovanni~C. Santia and Jake~Ryland Williams. 2018.
\newblock \href {https://aaai.org/ocs/index.php/ICWSM/ICWSM18/paper/view/17825}
  {{BuzzFace}: {A} news veracity dataset with facebook user commentary and
  egos}.
\newblock In \emph{Proceedings of the Twelfth International Conference on Web
  and Social Media, {ICWSM} 2018, Stanford, California, USA, June 25-28, 2018},
  pages 531--540. {AAAI} Press.

\bibitem[{Sathe et~al.(2020)Sathe, Ather, Le, Perry, and Park}]{SatheALPP20}
Aalok Sathe, Salar Ather, Tuan~Manh Le, Nathan Perry, and Joonsuk Park. 2020.
\newblock \href {https://aclanthology.org/2020.lrec-1.849/} {Automated
  fact-checking of claims from wikipedia}.
\newblock In \emph{Proceedings of The 12th Language Resources and Evaluation
  Conference, {LREC} 2020, Marseille, France, May 11-16, 2020}, pages
  6874--6882. European Language Resources Association.

\bibitem[{Schlichtkrull et~al.(2021)Schlichtkrull, Karpukhin, Oguz, Lewis, Yih,
  and Riedel}]{schlichtkrull2020}
Michael~Sejr Schlichtkrull, Vladimir Karpukhin, Barlas Oguz, Mike Lewis,
  Wen-tau Yih, and Sebastian Riedel. 2021.
\newblock \href {https://doi.org/10.18653/v1/2021.acl-long.529} {Joint
  verification and reranking for open fact checking over tables}.
\newblock In \emph{Proceedings of the 59th Annual Meeting of the Association
  for Computational Linguistics and the 11th International Joint Conference on
  Natural Language Processing (Volume 1: Long Papers)}, pages 6787--6799,
  Online. Association for Computational Linguistics.

\bibitem[{Schuster et~al.(2021)Schuster, Fisch, and Barzilay}]{vitaminc2021}
Tal Schuster, Adam Fisch, and Regina Barzilay. 2021.
\newblock \href {https://www.aclweb.org/anthology/2021.naacl-main.52} {Get your
  {Vitamin C}! robust fact verification with contrastive evidence}.
\newblock In \emph{Proceedings of the 2021 Conference of the North American
  Chapter of the Association for Computational Linguistics: Human Language
  Technologies}, pages 624--643, Online. Association for Computational
  Linguistics.

\bibitem[{Schuster et~al.(2020)Schuster, Schuster, Shah, and
  Barzilay}]{Schuster2020TheLO}
Tal Schuster, Roei Schuster, Darsh~J. Shah, and Regina Barzilay. 2020.
\newblock \href {https://doi.org/10.1162/coli_a_00380} {The limitations of
  stylometry for detecting machine-generated fake news}.
\newblock \emph{Computational Linguistics}, 46(2):499--510.

\bibitem[{Schuster et~al.(2019)Schuster, Shah, Yeo, Roberto Filizzola~Ortiz,
  Santus, and Barzilay}]{Schuster2019TowardsDF}
Tal Schuster, Darsh Shah, Yun Jie~Serene Yeo, Daniel Roberto Filizzola~Ortiz,
  Enrico Santus, and Regina Barzilay. 2019.
\newblock \href {https://doi.org/10.18653/v1/D19-1341} {Towards debiasing fact
  verification models}.
\newblock In \emph{Proceedings of the 2019 Conference on Empirical Methods in
  Natural Language Processing and the 9th International Joint Conference on
  Natural Language Processing (EMNLP-IJCNLP)}, pages 3419--3425, Hong Kong,
  China. Association for Computational Linguistics.

\bibitem[{Serrano and Smith(2019)}]{Serrano2019IsAI}
Sofia Serrano and Noah~A. Smith. 2019.
\newblock \href {https://doi.org/10.18653/v1/P19-1282} {Is attention
  interpretable?}
\newblock In \emph{Proceedings of the 57th Annual Meeting of the Association
  for Computational Linguistics}, pages 2931--2951, Florence, Italy.
  Association for Computational Linguistics.

\bibitem[{Shaar et~al.(2020)Shaar, Babulkov, Da~San~Martino, and
  Nakov}]{shaar-etal-2020-known}
Shaden Shaar, Nikolay Babulkov, Giovanni Da~San~Martino, and Preslav Nakov.
  2020.
\newblock \href {https://doi.org/10.18653/v1/2020.acl-main.332} {That is a
  known lie: Detecting previously fact-checked claims}.
\newblock In \emph{Proceedings of the 58th Annual Meeting of the Association
  for Computational Linguistics}, pages 3607--3618, Online. Association for
  Computational Linguistics.

\bibitem[{Shahi and Nandini(2020)}]{shahifakecovid}
Gautam~Kishore Shahi and Durgesh Nandini. 2020.
\newblock \href {http://workshop-proceedings.icwsm.org/pdf/2020_14.pdf}
  {Fake{C}ovid -- a multilingual cross-domain fact check news dataset for
  covid-19}.
\newblock In \emph{Workshop Proceedings of the 14th International {AAAI}
  {C}onference on {W}eb and {S}ocial {M}edia}.

\bibitem[{Sheng et~al.(2021)Sheng, Cao, Zhang, Li, and
  Zhong}]{sheng-etal-2021-article}
Qiang Sheng, Juan Cao, Xueyao Zhang, Xirong Li, and Lei Zhong. 2021.
\newblock \href {https://doi.org/10.18653/v1/2021.acl-long.425} {Article
  reranking by memory-enhanced key sentence matching for detecting previously
  fact-checked claims}.
\newblock In \emph{Proceedings of the 59th Annual Meeting of the Association
  for Computational Linguistics and the 11th International Joint Conference on
  Natural Language Processing (Volume 1: Long Papers)}, pages 5468--5481,
  Online. Association for Computational Linguistics.

\bibitem[{Shi and Weninger(2016)}]{ShiW16}
Baoxu Shi and Tim Weninger. 2016.
\newblock \href {https://doi.org/10.1016/j.knosys.2016.04.015} {Discriminative
  predicate path mining for fact checking in knowledge graphs}.
\newblock \emph{Knowl. Based Syst.}, 104:123--133.

\bibitem[{Shiralkar et~al.(2017)Shiralkar, Flammini, Menczer, and
  Ciampaglia}]{Shiralkar2017FindingSI}
Prashant Shiralkar, Alessandro Flammini, Filippo Menczer, and Giovanni~Luca
  Ciampaglia. 2017.
\newblock \href {https://doi.org/10.1109/ICDM.2017.105} {Finding streams in
  knowledge graphs to support fact checking}.
\newblock In \emph{2017 {IEEE} International Conference on Data Mining, {ICDM}
  2017, New Orleans, LA, USA, November 18-21, 2017}, pages 859--864. {IEEE}
  Computer Society.

\bibitem[{Shu et~al.(2019)Shu, Cui, Wang, Lee, and Liu}]{Shu2019dEFENDEF}
Kai Shu, Limeng Cui, Suhang Wang, Dongwon Lee, and Huan Liu. 2019.
\newblock \href {https://doi.org/10.1145/3292500.3330935} {{dEFEND}:
  Explainable fake news detection}.
\newblock In \emph{Proceedings of the 25th {ACM} {SIGKDD} International
  Conference on Knowledge Discovery {\&} Data Mining, {KDD} 2019, Anchorage,
  AK, USA, August 4-8, 2019}, pages 395--405. {ACM}.

\bibitem[{Shu et~al.(2020)Shu, Mahudeswaran, Wang, Lee, and
  Liu}]{Shu2018FakeNewsNetAD}
Kai Shu, Deepak Mahudeswaran, Suhang Wang, Dongwon Lee, and Huan Liu. 2020.
\newblock \href {https://doi.org/10.1089/big.2020.0062} {{FakeNewsNet}: {A}
  data repository with news content, social context, and spatiotemporal
  information for studying fake news on social media}.
\newblock \emph{Big Data}, 8(3):171--188.

\bibitem[{Shu et~al.(2017)Shu, Sliva, Wang, Tang, and Liu}]{shu2017fake}
Kai Shu, Amy Sliva, Suhang Wang, Jiliang Tang, and Huan Liu. 2017.
\newblock \href {https://doi.org/10.1145/3137597.3137600} {Fake news detection
  on social media: {A} data mining perspective}.
\newblock \emph{{SIGKDD} Explor.}, 19(1):22--36.

\bibitem[{da~Silva et~al.(2019)da~Silva, Vieira, and Garcia}]{SilvaVG19}
Fernando Cardoso~Durier da~Silva, Rafael Vieira, and Ana Cristina~Bicharra
  Garcia. 2019.
\newblock \href {http://hdl.handle.net/10125/59713} {Can machines learn to
  detect fake news? {A} survey focused on social media}.
\newblock In \emph{52nd Hawaii International Conference on System Sciences,
  {HICSS} 2019, Grand Wailea, Maui, Hawaii, USA, January 8-11, 2019}, pages
  1--8. ScholarSpace.

\bibitem[{Silverman(2014)}]{silverman2014verification}
Craig Silverman. 2014.
\newblock \emph{Verification Handbook: An Ultimate Guideline on Digital Age
  Sourcing for Emergency Coverage}.
\newblock European Journalism Centre.

\bibitem[{Socher et~al.(2013)Socher, Chen, Manning, and Ng}]{SocherCMN13}
Richard Socher, Danqi Chen, Christopher~D. Manning, and Andrew~Y. Ng. 2013.
\newblock \href
  {https://proceedings.neurips.cc/paper/2013/hash/b337e84de8752b27eda3a12363109e80-Abstract.html}
  {Reasoning with neural tensor networks for knowledge base completion}.
\newblock In \emph{Advances in Neural Information Processing Systems 26: 27th
  Annual Conference on Neural Information Processing Systems 2013. Proceedings
  of a meeting held December 5-8, 2013, Lake Tahoe, Nevada, United States},
  pages 926--934.

\bibitem[{Sunstein and Vermeule(2009)}]{sunstein2009conspiracy}
Cass~R Sunstein and Adrian Vermeule. 2009.
\newblock Conspiracy theories: Causes and cures.
\newblock \emph{Journal of Political Philosophy}, 17(2):202--227.

\bibitem[{Tai et~al.(2015)Tai, Socher, and Manning}]{tai2015improved}
Kai~Sheng Tai, Richard Socher, and Christopher~D. Manning. 2015.
\newblock \href {https://doi.org/10.3115/v1/P15-1150} {Improved semantic
  representations from tree-structured long short-term memory networks}.
\newblock In \emph{Proceedings of the 53rd Annual Meeting of the Association
  for Computational Linguistics and the 7th International Joint Conference on
  Natural Language Processing (Volume 1: Long Papers)}, pages 1556--1566,
  Beijing, China. Association for Computational Linguistics.

\bibitem[{Taylor(2003)}]{taylor2003munitions}
Philip~M. Taylor. 2003.
\newblock \emph{Munitions of the mind: A history of propaganda from the ancient
  world to the present era}, 3rd edition.
\newblock Manchester University Press.

\bibitem[{Thorne and Vlachos(2018)}]{Thorne2018AutomatedFC}
James Thorne and Andreas Vlachos. 2018.
\newblock \href {https://www.aclweb.org/anthology/C18-1283} {Automated fact
  checking: Task formulations, methods and future directions}.
\newblock In \emph{Proceedings of the 27th International Conference on
  Computational Linguistics}, pages 3346--3359, Santa Fe, New Mexico, USA.
  Association for Computational Linguistics.

\bibitem[{Thorne and Vlachos(2021)}]{Thorne2021ElasticWC}
James Thorne and Andreas Vlachos. 2021.
\newblock \href {https://www.aclweb.org/anthology/2021.eacl-main.82} {Elastic
  weight consolidation for better bias inoculation}.
\newblock In \emph{Proceedings of the 16th Conference of the European Chapter
  of the Association for Computational Linguistics: Main Volume}, pages
  957--964, Online. Association for Computational Linguistics.

\bibitem[{Thorne et~al.(2018{\natexlab{a}})Thorne, Vlachos, Christodoulopoulos,
  and Mittal}]{Thorne2018FEVERAL}
James Thorne, Andreas Vlachos, Christos Christodoulopoulos, and Arpit Mittal.
  2018{\natexlab{a}}.
\newblock \href {https://doi.org/10.18653/v1/N18-1074} {{FEVER}: a large-scale
  dataset for fact extraction and {VER}ification}.
\newblock In \emph{Proceedings of the 2018 Conference of the North {A}merican
  Chapter of the Association for Computational Linguistics: Human Language
  Technologies, Volume 1 (Long Papers)}, pages 809--819, New Orleans,
  Louisiana. Association for Computational Linguistics.

\bibitem[{Thorne et~al.(2018{\natexlab{b}})Thorne, Vlachos, Cocarascu,
  Christodoulopoulos, and Mittal}]{Thorne18Fact}
James Thorne, Andreas Vlachos, Oana Cocarascu, Christos Christodoulopoulos, and
  Arpit Mittal. 2018{\natexlab{b}}.
\newblock \href {https://doi.org/10.18653/v1/W18-5501} {The fact extraction and
  {VER}ification ({FEVER}) shared task}.
\newblock In \emph{Proceedings of the First Workshop on Fact Extraction and
  {VER}ification ({FEVER})}, pages 1--9, Brussels, Belgium. Association for
  Computational Linguistics.

\bibitem[{Thorne et~al.(2019)Thorne, Vlachos, Cocarascu, Christodoulopoulos,
  and Mittal}]{Thorne2019TheFS}
James Thorne, Andreas Vlachos, Oana Cocarascu, Christos Christodoulopoulos, and
  Arpit Mittal. 2019.
\newblock \href {https://doi.org/10.18653/v1/D19-6601} {The {FEVER}2.0 shared
  task}.
\newblock In \emph{Proceedings of the Second Workshop on Fact Extraction and
  VERification (FEVER)}, pages 1--6, Hong Kong, China. Association for
  Computational Linguistics.

\bibitem[{Tucker et~al.(2018)Tucker, Guess, Barber{\'a}, Vaccari, Siegel,
  Sanovich, Stukal, and Nyhan}]{Tucker2018SocialMP}
Joshua~A Tucker, Andrew Guess, Pablo Barber{\'a}, Cristian Vaccari, Alexandra
  Siegel, Sergey Sanovich, Denis Stukal, and Brendan Nyhan. 2018.
\newblock Social media, political polarization, and political disinformation: A
  review of the scientific literature.
\newblock \emph{Political polarization, and political disinformation: a review
  of the scientific literature (March 19, 2018)}.

\bibitem[{Uscinski(2015)}]{uscinski2015epistemology}
Joseph~E. Uscinski. 2015.
\newblock The epistemology of fact checking (is still na{\`\i}ve): Rejoinder to
  amazeen.
\newblock \emph{Critical Review}, 27(2):243--252.

\bibitem[{Uscinski and Butler(2013)}]{uscinski2013epistemology}
Joseph~E. Uscinski and Ryden~W. Butler. 2013.
\newblock The epistemology of fact checking.
\newblock \emph{Critical Review}, 25(2):162--180.

\bibitem[{Utama et~al.(2020{\natexlab{a}})Utama, Moosavi, and
  Gurevych}]{Utama2020MindTT}
Prasetya~Ajie Utama, Nafise~Sadat Moosavi, and Iryna Gurevych.
  2020{\natexlab{a}}.
\newblock \href {https://doi.org/10.18653/v1/2020.acl-main.770} {Mind the
  trade-off: Debiasing {NLU} models without degrading the in-distribution
  performance}.
\newblock In \emph{Proceedings of the 58th Annual Meeting of the Association
  for Computational Linguistics}, pages 8717--8729, Online. Association for
  Computational Linguistics.

\bibitem[{Utama et~al.(2020{\natexlab{b}})Utama, Moosavi, and
  Gurevych}]{Utama2020TowardsDN}
Prasetya~Ajie Utama, Nafise~Sadat Moosavi, and Iryna Gurevych.
  2020{\natexlab{b}}.
\newblock \href {https://doi.org/10.18653/v1/2020.emnlp-main.613} {Towards
  debiasing {NLU} models from unknown biases}.
\newblock In \emph{Proceedings of the 2020 Conference on Empirical Methods in
  Natural Language Processing (EMNLP)}, pages 7597--7610, Online. Association
  for Computational Linguistics.

\bibitem[{Vlachos and Riedel(2014)}]{Vlachos2014FactCT}
Andreas Vlachos and Sebastian Riedel. 2014.
\newblock \href {https://doi.org/10.3115/v1/W14-2508} {Fact checking: Task
  definition and dataset construction}.
\newblock In \emph{Proceedings of the {ACL} 2014 Workshop on Language
  Technologies and Computational Social Science}, pages 18--22, Baltimore, MD,
  USA. Association for Computational Linguistics.

\bibitem[{Vlachos and Riedel(2015)}]{vlachos-riedel-2015-identification}
Andreas Vlachos and Sebastian Riedel. 2015.
\newblock \href {https://doi.org/10.18653/v1/D15-1312} {Identification and
  verification of simple claims about statistical properties}.
\newblock In \emph{Proceedings of the 2015 Conference on Empirical Methods in
  Natural Language Processing}, pages 2596--2601, Lisbon, Portugal. Association
  for Computational Linguistics.

\bibitem[{Vo and Lee(2020)}]{Vo2020WhereAT}
Nguyen Vo and Kyumin Lee. 2020.
\newblock \href {https://doi.org/10.18653/v1/2020.emnlp-main.621} {Where are
  the facts? searching for fact-checked information to alleviate the spread of
  fake news}.
\newblock In \emph{Proceedings of the 2020 Conference on Empirical Methods in
  Natural Language Processing (EMNLP)}, pages 7717--7731, Online. Association
  for Computational Linguistics.

\bibitem[{Volkova et~al.(2017)Volkova, Shaffer, Jang, and
  Hodas}]{Volkova2017SeparatingFF}
Svitlana Volkova, Kyle Shaffer, Jin~Yea Jang, and Nathan Hodas. 2017.
\newblock \href {https://doi.org/10.18653/v1/P17-2102} {Separating facts from
  fiction: Linguistic models to classify suspicious and trusted news posts on
  {T}witter}.
\newblock In \emph{Proceedings of the 55th Annual Meeting of the Association
  for Computational Linguistics (Volume 2: Short Papers)}, pages 647--653,
  Vancouver, Canada. Association for Computational Linguistics.

\bibitem[{Wadden et~al.(2020)Wadden, Lin, Lo, Wang, van Zuylen, Cohan, and
  Hajishirzi}]{Wadden2020FactOF}
David Wadden, Shanchuan Lin, Kyle Lo, Lucy~Lu Wang, Madeleine van Zuylen, Arman
  Cohan, and Hannaneh Hajishirzi. 2020.
\newblock \href {https://doi.org/10.18653/v1/2020.emnlp-main.609} {{Fact or
  Fiction}: Verifying scientific claims}.
\newblock In \emph{Proceedings of the 2020 Conference on Empirical Methods in
  Natural Language Processing (EMNLP)}, pages 7534--7550, Online. Association
  for Computational Linguistics.

\bibitem[{Wang et~al.(2021)Wang, Mahajan, Danilevsky, and
  Rosenthal}]{wang2021semeval}
Nancy Xin~Ru Wang, Diwakar Mahajan, Marina Danilevsky, and Sara Rosenthal.
  2021.
\newblock \href {https://doi.org/10.18653/v1/2021.semeval-1.39} {{SemEval-2021}
  task 9: Fact verification and evidence finding for tabular data in scientific
  documents {(SEM-TAB-FACTS)}}.
\newblock In \emph{Proceedings of the 15th International Workshop on Semantic
  Evaluation, SemEval@ACL/IJCNLP 2021, Virtual Event / Bangkok, Thailand,
  August 5-6, 2021}, pages 317--326. Association for Computational Linguistics.

\bibitem[{Wang(2017)}]{Wang2017LiarLP}
William~Yang Wang. 2017.
\newblock \href {https://doi.org/10.18653/v1/P17-2067} {{``}{Liar, Liar Pants
  on Fire{''}}: A new benchmark dataset for fake news detection}.
\newblock In \emph{Proceedings of the 55th Annual Meeting of the Association
  for Computational Linguistics (Volume 2: Short Papers)}, pages 422--426,
  Vancouver, Canada. Association for Computational Linguistics.

\bibitem[{Wiegreffe and Pinter(2019)}]{wiegreffe2019attention}
Sarah Wiegreffe and Yuval Pinter. 2019.
\newblock \href {https://doi.org/10.18653/v1/D19-1002} {Attention is not not
  explanation}.
\newblock In \emph{Proceedings of the 2019 Conference on Empirical Methods in
  Natural Language Processing and the 9th International Joint Conference on
  Natural Language Processing (EMNLP-IJCNLP)}, pages 11--20, Hong Kong, China.
  Association for Computational Linguistics.

\bibitem[{Williams et~al.(2018)Williams, Nangia, and Bowman}]{Williams2018ABC}
Adina Williams, Nikita Nangia, and Samuel Bowman. 2018.
\newblock \href {https://doi.org/10.18653/v1/N18-1101} {A broad-coverage
  challenge corpus for sentence understanding through inference}.
\newblock In \emph{Proceedings of the 2018 Conference of the North {A}merican
  Chapter of the Association for Computational Linguistics: Human Language
  Technologies, Volume 1 (Long Papers)}, pages 1112--1122, New Orleans,
  Louisiana. Association for Computational Linguistics.

\bibitem[{Wu et~al.(2020{\natexlab{a}})Wu, Rao, Yang, Wang, and
  Nazir}]{wu2020evidenceaware}
Lianwei Wu, Yuan Rao, Xiong Yang, Wanzhen Wang, and Ambreen Nazir.
  2020{\natexlab{a}}.
\newblock \href {https://doi.org/10.24963/ijcai.2020/193} {Evidence-aware
  hierarchical interactive attention networks for explainable claim
  verification}.
\newblock In \emph{Proceedings of the Twenty-Ninth International Joint
  Conference on Artificial Intelligence, {IJCAI-20}}, pages 1388--1394.
  International Joint Conferences on Artificial Intelligence Organization.
\newblock Main track.

\bibitem[{Wu et~al.(2020{\natexlab{b}})Wu, Rao, Zhao, Liang, and
  Nazir}]{Wu2020DTCADT}
Lianwei Wu, Yuan Rao, Yongqiang Zhao, Hao Liang, and Ambreen Nazir.
  2020{\natexlab{b}}.
\newblock \href {https://doi.org/10.18653/v1/2020.acl-main.97} {{DTCA}:
  Decision tree-based co-attention networks for explainable claim
  verification}.
\newblock In \emph{Proceedings of the 58th Annual Meeting of the Association
  for Computational Linguistics}, pages 1024--1035, Online. Association for
  Computational Linguistics.

\bibitem[{Yang et~al.(2019)Yang, Pentyala, Mohseni, Du, Yuan, Linder, Ragan,
  Ji, and Hu}]{Yang2019XFakeEF}
Fan Yang, Shiva~K. Pentyala, Sina Mohseni, Mengnan Du, Hao Yuan, Rhema Linder,
  Eric~D. Ragan, Shuiwang Ji, and Xia~(Ben) Hu. 2019.
\newblock \href {https://doi.org/10.1145/3308558.3314119} {{XFake}: Explainable
  fake news detector with visualizations}.
\newblock In \emph{The World Wide Web Conference, {WWW} 2019, San Francisco,
  CA, USA, May 13-17, 2019}, pages 3600--3604. {ACM}.

\bibitem[{Yang et~al.(2020{\natexlab{a}})Yang, Lyu, Tian, Liu, Liu, and
  Zhang}]{yang2020rumor}
Xiaoyu Yang, Yuefei Lyu, Tian Tian, Yifei Liu, Yudong Liu, and Xi~Zhang.
  2020{\natexlab{a}}.
\newblock \href {https://doi.org/10.24963/ijcai.2020/197} {Rumor detection on
  social media with graph structured adversarial learning}.
\newblock In \emph{Proceedings of the Twenty-Ninth International Joint
  Conference on Artificial Intelligence, {IJCAI} 2020}, pages 1417--1423.
  ijcai.org.

\bibitem[{Yang et~al.(2020{\natexlab{b}})Yang, Nie, Feng, Liu, Chen, and
  Zhu}]{yang-etal-2020-program}
Xiaoyu Yang, Feng Nie, Yufei Feng, Quan Liu, Zhigang Chen, and Xiaodan Zhu.
  2020{\natexlab{b}}.
\newblock \href {https://doi.org/10.18653/v1/2020.emnlp-main.628} {Program
  enhanced fact verification with verbalization and graph attention network}.
\newblock In \emph{Proceedings of the 2020 Conference on Empirical Methods in
  Natural Language Processing (EMNLP)}, pages 7810--7825, Online. Association
  for Computational Linguistics.

\bibitem[{Yeh et~al.(2019)Yeh, Hsieh, Suggala, Inouye, and
  Ravikumar}]{yeh2019infidelity}
Chih{-}Kuan Yeh, Cheng{-}Yu Hsieh, Arun~Sai Suggala, David~I. Inouye, and
  Pradeep Ravikumar. 2019.
\newblock \href
  {https://proceedings.neurips.cc/paper/2019/hash/a7471fdc77b3435276507cc8f2dc2569-Abstract.html}
  {On the (in)fidelity and sensitivity of explanations}.
\newblock In \emph{Advances in Neural Information Processing Systems 32: Annual
  Conference on Neural Information Processing Systems 2019, NeurIPS 2019,
  December 8-14, 2019, Vancouver, BC, Canada}, pages 10965--10976.

\bibitem[{Yin and Roth(2018)}]{Riloff18}
Wenpeng Yin and Dan Roth. 2018.
\newblock \href {https://doi.org/10.18653/v1/d18-1010} {{TwoWingOS}: {A}
  two-wing optimization strategy for evidential claim verification}.
\newblock In \emph{Proceedings of the 2018 Conference on Empirical Methods in
  Natural Language Processing, Brussels, Belgium, October 31 - November 4,
  2018}, pages 105--114. Association for Computational Linguistics.

\bibitem[{Yoneda et~al.(2018)Yoneda, Mitchell, Welbl, Stenetorp, and
  Riedel}]{yoneda-etal-2018-ucl}
Takuma Yoneda, Jeff Mitchell, Johannes Welbl, Pontus Stenetorp, and Sebastian
  Riedel. 2018.
\newblock \href {https://doi.org/10.18653/v1/W18-5515} {{UCL} machine reading
  group: Four factor framework for fact finding ({H}exa{F})}.
\newblock In \emph{Proceedings of the First Workshop on Fact Extraction and
  {VER}ification ({FEVER})}, pages 97--102, Brussels, Belgium. Association for
  Computational Linguistics.

\bibitem[{Yu et~al.(2019)Yu, Chang, Zhang, and Jaakkola}]{yu2019rethinking}
Mo~Yu, Shiyu Chang, Yang Zhang, and Tommi Jaakkola. 2019.
\newblock \href {https://doi.org/10.18653/v1/D19-1420} {Rethinking cooperative
  rationalization: Introspective extraction and complement control}.
\newblock In \emph{Proceedings of the 2019 Conference on Empirical Methods in
  Natural Language Processing and the 9th International Joint Conference on
  Natural Language Processing (EMNLP-IJCNLP)}, pages 4094--4103, Hong Kong,
  China. Association for Computational Linguistics.

\bibitem[{Zeichner et~al.(2012)Zeichner, Berant, and Dagan}]{zeichner2012}
Naomi Zeichner, Jonathan Berant, and Ido Dagan. 2012.
\newblock \href {https://aclanthology.org/P12-2031} {Crowdsourcing
  inference-rule evaluation}.
\newblock In \emph{Proceedings of the 50th Annual Meeting of the Association
  for Computational Linguistics (Volume 2: Short Papers)}, pages 156--160, Jeju
  Island, Korea. Association for Computational Linguistics.

\bibitem[{Zellers et~al.(2019)Zellers, Holtzman, Rashkin, Bisk, Farhadi,
  Roesner, and Choi}]{Zellers2019DefendingAN}
Rowan Zellers, Ari Holtzman, Hannah Rashkin, Yonatan Bisk, Ali Farhadi,
  Franziska Roesner, and Yejin Choi. 2019.
\newblock \href
  {https://proceedings.neurips.cc/paper/2019/hash/3e9f0fc9b2f89e043bc6233994dfcf76-Abstract.html}
  {Defending against neural fake news}.
\newblock In \emph{Advances in Neural Information Processing Systems 32: Annual
  Conference on Neural Information Processing Systems 2019, NeurIPS 2019,
  December 8-14, 2019, Vancouver, BC, Canada}, pages 9051--9062.

\bibitem[{Zhang et~al.(2018)Zhang, Shang, Geng, Lai, Li, Zhu, Amin, and
  Wang}]{Zhang2018FauxBusterAC}
Daniel~Yue Zhang, Lanyu Shang, Biao Geng, Shuyue Lai, Ke~Li, Hongmin Zhu, Md.
  Tanvir~Al Amin, and Dong Wang. 2018.
\newblock \href {https://doi.org/10.1109/BigData.2018.8622344} {{FauxBuster}:
  {A} content-free fauxtography detector using social media comments}.
\newblock In \emph{{IEEE} International Conference on Big Data, Big Data 2018,
  Seattle, WA, USA, December 10-13, 2018}, pages 891--900. {IEEE}.

\bibitem[{Zhang et~al.(2020)Zhang, Deng, Ma, and Lam}]{Zhang2020AnswerFactFC}
Wenxuan Zhang, Yang Deng, Jing Ma, and Wai Lam. 2020.
\newblock \href {https://doi.org/10.18653/v1/2020.emnlp-main.188}
  {{A}nswer{F}act: Fact checking in product question answering}.
\newblock In \emph{Proceedings of the 2020 Conference on Empirical Methods in
  Natural Language Processing (EMNLP)}, pages 2407--2417, Online. Association
  for Computational Linguistics.

\bibitem[{Zhang et~al.(2021)Zhang, Cao, Li, Sheng, Zhong, and
  Shu}]{zhang2021mining}
Xueyao Zhang, Juan Cao, Xirong Li, Qiang Sheng, Lei Zhong, and Kai Shu. 2021.
\newblock \href {https://doi.org/10.1145/3442381.3450004} {Mining dual emotion
  for fake news detection}.
\newblock In \emph{{WWW} '21: The Web Conference 2021, Virtual Event /
  Ljubljana, Slovenia, April 19-23, 2021}, pages 3465--3476. {ACM} / {IW3C2}.

\bibitem[{Zhang et~al.(2019)Zhang, Ives, and Roth}]{zhang-etal-2019-evidence}
Yi~Zhang, Zachary Ives, and Dan Roth. 2019.
\newblock \href {https://doi.org/10.18653/v1/P19-1040} {Evidence-based
  trustworthiness}.
\newblock In \emph{Proceedings of the 57th Annual Meeting of the Association
  for Computational Linguistics}, pages 413--423, Florence, Italy. Association
  for Computational Linguistics.

\bibitem[{Zhong et~al.(2020{\natexlab{a}})Zhong, Tang, Feng, Duan, Zhou, Gong,
  Shou, Jiang, Wang, and Yin}]{zhong-etal-2020-logicalfactchecker}
Wanjun Zhong, Duyu Tang, Zhangyin Feng, Nan Duan, Ming Zhou, Ming Gong, Linjun
  Shou, Daxin Jiang, Jiahai Wang, and Jian Yin. 2020{\natexlab{a}}.
\newblock \href {https://doi.org/10.18653/v1/2020.acl-main.539}
  {{L}ogical{F}act{C}hecker: Leveraging logical operations for fact checking
  with graph module network}.
\newblock In \emph{Proceedings of the 58th Annual Meeting of the Association
  for Computational Linguistics}, pages 6053--6065, Online. Association for
  Computational Linguistics.

\bibitem[{Zhong et~al.(2020{\natexlab{b}})Zhong, Xu, Tang, Xu, Duan, Zhou,
  Wang, and Yin}]{Zhong2020ReasoningOS}
Wanjun Zhong, Jingjing Xu, Duyu Tang, Zenan Xu, Nan Duan, Ming Zhou, Jiahai
  Wang, and Jian Yin. 2020{\natexlab{b}}.
\newblock \href {https://doi.org/10.18653/v1/2020.acl-main.549} {Reasoning over
  semantic-level graph for fact checking}.
\newblock In \emph{Proceedings of the 58th Annual Meeting of the Association
  for Computational Linguistics}, pages 6170--6180, Online. Association for
  Computational Linguistics.

\bibitem[{Zhou et~al.(2019)Zhou, Han, Yang, Liu, Wang, Li, and
  Sun}]{Zhou2019GEARGE}
Jie Zhou, Xu~Han, Cheng Yang, Zhiyuan Liu, Lifeng Wang, Changcheng Li, and
  Maosong Sun. 2019.
\newblock \href {https://doi.org/10.18653/v1/P19-1085} {{GEAR}: Graph-based
  evidence aggregating and reasoning for fact verification}.
\newblock In \emph{Proceedings of the 57th Annual Meeting of the Association
  for Computational Linguistics}, pages 892--901, Florence, Italy. Association
  for Computational Linguistics.

\bibitem[{Zhou et~al.(2020)Zhou, Jain, Phoha, and Zafarani}]{zhou2020fakenews}
Xinyi Zhou, Atishay Jain, Vir~V. Phoha, and Reza Zafarani. 2020.
\newblock \href {https://doi.org/10.1145/3377478} {Fake news early detection: A
  theory-driven model}.
\newblock \emph{Digital Threats: Research and Practice}, 1(2).

\bibitem[{Zhou and Zafarani(2020)}]{Zhou20}
Xinyi Zhou and Reza Zafarani. 2020.
\newblock \href {https://doi.org/10.1145/3395046} {A survey of fake news:
  Fundamental theories, detection methods, and opportunities}.
\newblock \emph{{ACM} Computing Surveys}, 53(5):109:1--109:40.

\bibitem[{Zlatkova et~al.(2019)Zlatkova, Nakov, and
  Koychev}]{Zlatkova2019FactCheckingMF}
Dimitrina Zlatkova, Preslav Nakov, and Ivan Koychev. 2019.
\newblock \href {https://doi.org/10.18653/v1/D19-1216} {Fact-checking meets
  fauxtography: Verifying claims about images}.
\newblock In \emph{Proceedings of the 2019 Conference on Empirical Methods in
  Natural Language Processing and the 9th International Joint Conference on
  Natural Language Processing (EMNLP-IJCNLP)}, pages 2099--2108, Hong Kong,
  China. Association for Computational Linguistics.

\bibitem[{Zubiaga et~al.(2018)Zubiaga, Aker, Bontcheva, Liakata, and
  Procter}]{ZubiagaABLP18}
Arkaitz Zubiaga, Ahmet Aker, Kalina Bontcheva, Maria Liakata, and Rob Procter.
  2018.
\newblock \href {https://doi.org/10.1145/3161603} {Detection and resolution of
  rumours in social media: {A} survey}.
\newblock \emph{{ACM} Computing Surveys}, 51(2):32:1--32:36.

\bibitem[{Zubiaga et~al.(2016)Zubiaga, Liakata, Procter, Wong Sak~Hoi, and
  Tolmie}]{zubiaga2016analysing}
Arkaitz Zubiaga, Maria Liakata, Rob Procter, Geraldine Wong Sak~Hoi, and Peter
  Tolmie. 2016.
\newblock Analysing how people orient to and spread rumours in social media by
  looking at conversational threads.
\newblock \emph{PloS one}, 11(3):e0150989.

\bibitem[{Zuo et~al.(2018)Zuo, Karakas, and Banerjee}]{Zuo2018AHR}
Chaoyuan Zuo, Ayla Karakas, and Ritwik Banerjee. 2018.
\newblock \href {http://ceur-ws.org/Vol-2125/paper\_143.pdf} {A hybrid
  recognition system for check-worthy claims using heuristics and supervised
  learning}.
\newblock In \emph{Working Notes of {CLEF} 2018 - Conference and Labs of the
  Evaluation Forum, Avignon, France, September 10-14, 2018}, volume 2125 of
  \emph{{CEUR} Workshop Proceedings}. CEUR-WS.org.

\end{thebibliography}
\bibliographystyle{acl_natbib}

\end{document}